\documentclass[acmsmall]{acmart}
\usepackage{tikz}
\usepackage{graphicx}
\usepackage[edges]{forest}
\usepackage{subcaption}

\usepackage{pifont}
\usepackage{bbm}
\newcommand{\cmark}{\textcolor{green!60!black}{\ding{51}}} 
\newcommand{\xmark}{\textcolor{red}{\ding{55}}} 

\usepackage[table]{xcolor}

\newcommand{\revised}[1]{{\color{black}{#1}}}

\newcommand{\commentisunseen}{0}

\definecolor{myred}{RGB}{246, 204, 204}
\definecolor{myblue}{RGB}{164, 194, 244}
\definecolor{mygray}{RGB}{185, 185, 185}
\definecolor{myyellow}{RGB}{255, 221, 179}
\definecolor{mypurple}{RGB}{213, 194, 234}   
\definecolor{mycyan}{RGB}{179, 226, 226}     
\definecolor{myorange}{RGB}{255, 196, 153}  
\definecolor{mydeepgray}{RGB}{150, 150, 150} 
\definecolor{mypink}{RGB}{250, 204, 213}      
\definecolor{myolive}{RGB}{212, 220, 182}    
\definecolor{myblueline}{RGB}{87, 127, 185}
\definecolor{bluelight1}{RGB}{185, 211, 237}
\definecolor{bluelight2}{RGB}{213, 222, 239}
\definecolor{mygreen}{RGB}{168, 209, 201}
\definecolor{greenlight}{RGB}{220, 235, 234}
\definecolor{hidden-draw}{RGB}{177, 177, 177}

\definecolor{myultralightgray}{RGB}{240, 240, 240}  
\definecolor{myultralightpink}{RGB}{252, 228, 232}  
\definecolor{myultralightblue}{RGB}{220, 230, 248}  
\definecolor{myultralightgreen}{RGB}{225, 240, 225}  
\definecolor{myultralightyellow}{RGB}{255, 240, 210} 
\definecolor{myultralightpurple}{RGB}{235, 228, 242} 
\definecolor{myultralightbrown}{RGB}{240, 225, 210} 

\newcommand{\commentp}[1]{
}

\newcommand{\fc}[1]{
\ifthenelse{\equal{\commentisunseen}{0}}{
{\color{blue}#1}}
{#1}
}
\newcommand{\FC}[1]{
\ifthenelse{\equal{\commentisunseen}{0}}{
{\color{blue}FC: #1}}
{}
}

\AtBeginDocument{
  }

\setcopyright{acmlicensed}
\copyrightyear{0000}
\acmYear{0000}
\acmDOI{0000000.0000000}

\acmJournal{JACM}
\acmVolume{00}
\acmNumber{0}
\acmArticle{000}
\acmMonth{0}

\begin{document}

\title{Model Merging in LLMs, MLLMs, and Beyond: Methods, Theories, Applications and Opportunities}

\author{Enneng Yang}
\email{ennengyang@gmail.com}
\affiliation{
  \institution{Shenzhen Campus of Sun Yat-sen University, China; Northeastern University}
  \country{China}
}

\author{Li Shen}
\authornote{Corresponding author.}
\email{shenli6@mail.sysu.edu.cn}
\affiliation{
  \institution{Shenzhen Campus of Sun Yat-sen University}
  \country{China}
}

\author{Guibing Guo}
\authornotemark[1]
\email{guogb@swc.neu.edu.cn}
\affiliation{
  \institution{Northeastern University}
  \country{China}
}

\author{Xingwei Wang}
\email{wangxw@mail.neu.edu.cn}
\affiliation{
  \institution{Northeastern University}
  \country{China}
}

\author{Xiaochun Cao}
\email{caoxiaochun@mail.sysu.edu.cn}
\affiliation{
  \institution{Shenzhen Campus of Sun Yat-sen University}
  \country{China}
}

\author{Jie Zhang}
\email{zhangj@ntu.edu.sg}
\affiliation{
  \institution{Nanyang Technological University}
  \country{Singapore}
}

\author{Dacheng Tao}
\email{dacheng.tao@ntu.edu.sg}
\affiliation{
  \institution{Nanyang Technological University}
  \country{Singapore}
}

\begin{abstract}
Model merging is an efficient empowerment technique in the machine learning community that does not require the collection of raw training data and does not require expensive computation. As model merging becomes increasingly prevalent across various fields, it is crucial to understand the available model merging techniques comprehensively. However, there is a significant gap in the literature regarding a systematic and thorough review of these techniques. This survey provides a comprehensive overview of model merging methods and theories, their applications in various domains and settings, and future research directions. Specifically, we first propose a new taxonomic approach that exhaustively discusses existing model merging methods. Secondly, we discuss the application of model merging techniques in large language models, multimodal large language models, and more than ten machine learning subfields, including continual learning, multi-task learning, few-shot learning, etc. Finally, we highlight the remaining challenges of model merging and discuss future research directions. A comprehensive list of papers about model merging is available at \textit{\url{https://github.com/EnnengYang/Awesome-Model-Merging-Methods-Theories-Applications}}.
\end{abstract}



\maketitle

\section{Introduction}
\label{sec:introduction}

\begin{figure*}[t]
\centering  
\includegraphics[width=1.0\textwidth]{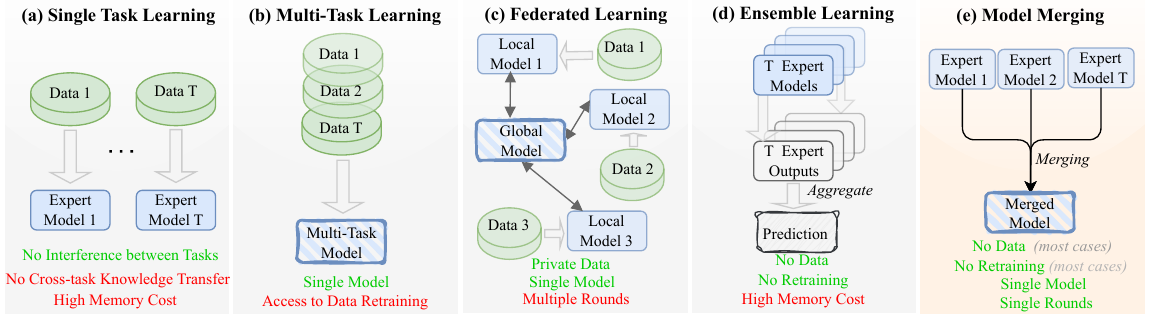}
\vspace{-20pt}
\caption{
\revised{
Schematic comparison of five machine learning paradigms (adapted from \cite{RegMean_ICLR2023}): 
(a) Single Task Learning, (b) Multi-Task Learning, (c) Federated Learning, (d) Ensemble Learning, (e) Model Merging.
Key attributes are highlighted in red and green.
}
}
\vspace{-15pt}
\label{fig:same_architectures}
\end{figure*}

Model merging, also known as model fusion, is an effective technique that merges the parameters of multiple expert models with different capabilities to build a universal model without needing access to the original training data or expensive computation. The concepts most relevant to model merging are multi-task learning~\cite{SharedBottom_1997,mtlsurvey_tpami2021}, federated learning~\cite{li2020federated}, and ensemble learning~\cite{ensemblelearning_book2002,sagi2018ensemblesurvey}, as all aim to facilitate knowledge fusion and transfer. 
\revised{
Specifically, as shown in Figure~\ref{fig:same_architectures}, single-task learning trains a separate expert model for each task; while this avoids interference across tasks, it hinders knowledge transfer and incurs high storage costs. Multi-task learning maintains a single shared model, but requires collecting data from all tasks to jointly update the model parameters. Federated learning trains local models on clients’ private data and then aggregates them on a central server to obtain a global model, but this requires multiple rounds of communication between the server and clients. Ensemble learning aggregates the outputs of multiple expert models; however, it must store all expert models, leading again to substantial storage overhead. In contrast, model merging directly aggregates the \textit{parameters} of several well-trained expert models to fuse their knowledge, making it highly efficient and inexpensive. These properties render model merging a particularly attractive paradigm for knowledge integration.
}

Although model merging is a relatively young topic, it is evolving rapidly and has already found applications in several domains. For example, in foundation models, models fine-tuned by different downstream tasks are merged to enhance the capabilities of large language models, and visual generative models with different styles are merged to create a new model with mixed-style capabilities. In particular, the number of pre-trained and fine-tuned checkpoints in the machine learning community has grown exponentially in recent years, including open-source repositories such as Huggingface~\cite{wolf2019huggingface}, torchvision~\cite{marcel2010torchvision}, and timm~\cite{rw2019timm}, making it easy for users to obtain well-trained expert models of varying abilities. These rich model repositories further promote the rapid development of the model merging direction.

As model merging becomes increasingly popular in various areas of the machine learning community, it is crucial to have a comprehensive understanding of the advantages and limitations of existing model merging techniques and their applications across different domains and settings. Although some efforts have been made by the community~\cite{modelfusion_survey2023,zheng2023learn,tangFusionBenchComprehensiveBenchmark2024,arceemergekit2024}, there are still large gaps to be filled. More specifically, MergeKit~\cite{arceemergekit2024} and FusionBench~\cite{tangFusionBenchComprehensiveBenchmark2024} are benchmark works in which only seven representative methods are discussed in MergeKit, and eight merging methods are discussed in FusionBench. Additionally, \citet{zheng2023learn} discuss the topic of "learning from models" and it only mentions model merging as a subsection (single page only) in the whole paper. The most related work to the "model merging" topic is \cite{modelfusion_survey2023}, but in terms of application, it only discusses model merging in three scenarios: federated learning, fine-tuning, and distillation. It also ignores a lot of recently published articles due to the rapid evolution of the model merging direction.
To address these gaps, this survey aims to elucidate the methods, theories, applications, and future trends in model merging direction, providing a comprehensive classification of relevant approaches. In particular, this paper enhances the comprehensive understanding of model merging by covering three main aspects:

\textbf{First, how are existing model merging methods classified}?
We first propose a new taxonomy in Figure~\ref{fig:taxonomy} (upper part) that divides existing model merging methods into two phases (\S \ref{sec:methods}): pre-merging and during-merging. 
(i) \textit{Pre-merging methods} aim to create better conditions for merging. It is further divided into using linearized fine-tuning to achieve weight space and input space disentanglement or subspace fine-tuning schemes that reduce parameter conflicts, performing architectural transformations to convert heterogeneous models into homogeneous models, and aligning weights to place them in the same basin.
(ii) \textit{During-merging methods} focus on designing sophisticated techniques to merge multiple models into one. These methods address task conflict and interference problems when merging models. They can be further divided into basic merging methods that perform the simplest parameter merging strategy; weighted merging methods that merge multiple models according to the importance calculated by specific rules; subspace merging methods that project multiple models into sparse or low-rank subspaces for merging; \revised{optimization-based methods leverage model-specific properties to define optimization objectives that minimize information loss};  routing-based methods that dynamically merge models according to input samples during inference; and the post-calibration-based method that corrects the merged model. In addition to these methods, we also discuss the theoretical or empirical analysis of model merging  (\S \ref{sec:theory}).

\tikzstyle{my-box}=[
    rectangle,
    draw=hidden-draw,
    rounded corners,
    text opacity=1,
    minimum height=1.5em,
    minimum width=5em,
    inner sep=2pt,
    align=center,
    fill opacity=.5
]
\tikzstyle{utilization_leaf}=[my-box, minimum height=1.5em,
    fill=bluelight2!100, text=black, align=left,font=\scriptsize,
    inner xsep=2pt,
    inner ysep=4pt
]
\tikzstyle{evolution_leaf}=[my-box, minimum height=1.5em,
    fill=greenlight, text=black, align=left,font=\scriptsize,
    inner xsep=2pt,
    inner ysep=4pt
]
\begin{figure*}[t]
    \centering
    \resizebox{\textwidth}{!}{
        \begin{forest}
            forked edges,
            for tree={
              grow=east,
              reversed=true,
              anchor=base west,
              parent anchor=east,
              child anchor=west,
              base=left,
              font=\small,
              rectangle,
              draw,
              rounded corners,
              align=left,
              minimum width=4em,
              edge+={darkgray, line width=1pt},
              s sep=3pt,
              inner xsep=2pt,
              inner ysep=3pt,
              ver/.style={rotate=90, child anchor=north, parent anchor=south, anchor=center}
            },
            [
                {Model Merging: Methods, Theories, Applications}, ver, 
                color=hidden-draw, fill=mygray!100, 
                text width=18.5em, 
                text=black
                [
                    Methods (\S \ref{sec:methods}), fill=myred!80, text width=9.0em, text=black
                    [
                        Pre-Merging Mehtods (\S\ref{subsec:beforemerging}),  fill=myred!60,  text width=13em, text=black
                            [
                               {
                               Merger-Friendly Fine-Tuning (\S\ref{subsubsec:linearization}) \\
                               Architecture Transformation (\S\ref{subsubsec:artchtecture})\\
                               Weight Re-basin or Alignment (\S\ref{subsubsec:alignment})
                               },
                                color=hidden-draw, fill=myred!40,  text width=21.5em, text=black
                            ]
                    ]
                    [
                        During-Merging Methods (\S \ref{subsec:duringmerging}),  fill=myred!60, text width=13em, text=black
                        [
                            {
                            Basic Merging Methods  (\S \ref{subsubsec:basic}) \\
                            Weighted-based Merging Methods  (\S \ref{subsubsec:weighted}) \\
                            Subspace-based Merging Methods  (\S \ref{subsubsec:subspace}) \\
                            Optimization-based Merging Methods  (\S \ref{subsubsec:optimization}) \\
                            Routing-based Merging Methods   (\S \ref{subsubsec:routing}) \\
                            Post-calibration-based Merging Methods (\S \ref{subsubsec:postmerging})
                            },
                            color=hidden-draw, fill=myred!40,  text width=21.5em, text=black
                        ]
                    ]
                ]
                [
                   {Theories and\\ Experiments (\S \ref{sec:theory})},  fill=mypurple!80, text width=9.0em, text=black
                   [
                        Theoretical Analysis (\S \ref{subsec:theory}), fill=mypurple!60, text width=13em, text=black
                   ]
                   [
                        Empirical Comparisons (\S \ref{subsec:EmpiricalComparisons}), fill=mypurple!60, text width=13em, text=black
                   ]
                ]
                [
                   Applications (\S \ref{sec:application_lm} \& \S \ref{sec:application_ml}), fill=myblue!80, text width=9.0em, text=black
                       [
                            Large Language Models (\S \ref{subsec:llms}), fill=myblue!60, text width=16em, text=black
                            [
                                {
                                Human Value Alignment for LLMs (\S \ref{subsubsec:llm_Alignment}) \\
                                Detoxification of LLMs (\S \ref{subsubsec:llm_Detoxifcation}) \\
                                Knowledge Unlearning of LLMs (\S \ref{subsubsec:llm_Unlearning}) \\
                                Faster Training of LLMs (\S \ref{subsubsec:llm_Faster}) \\
                                Faster Inference of LLMs (\S \ref{subsubsec:llm_FasterInference}) \\
                                Faster Reasoning of LLMs (\S \ref{subsubsec:llm_FasterReasoning}) \\
                                Combine the Capabilities of Expert LLMs (\S \ref{subsubsec:llm_CombineExpert})
                                }, 
                                color=hidden-draw, fill=myblue!40, text width=18.5em, text=black
                           ] 
                       ]
                       [
                           Multimodal Large Language Models (\S\ref{subsec:multimodal}), fill=myblue!60, text width=16em, text=black
                           [
                             {
                             Multimodal Fusion (\S\ref{subsubsec:multimodalfusion}) \\
                             Cross-modal Knowledge Transfer (\S\ref{subsubsec:cross-modal})
                             }, 
                             color=hidden-draw, fill=myblue!40, text width=18.5em, text=black
                           ]
                       ]
                       [
                              Visual Generative Models (\S \ref{subsec:generative}), fill=myblue!60, text width=16em, text=black
                            [
                                {
                                Style Mixing (\S \ref{subsubsec:generative_stylemix}) \\
                                Reducing Training Cost (\S \ref{subsubsec:generative_cost})
                                }, 
                                color=hidden-draw, fill=myblue!40, text width=18.5em, text=black
                           ] 
                       ]
                       [
                           Continual Learning (\S\ref{subsec:continual}),  fill=myyellow!60, text width=14em, text=black
                           [
                             Mitigate Catastrophic Forgetting (\S \ref{subsubsec:forgetting}), 
                             color=hidden-draw, fill=myyellow!40, text width=20.5em, text=black
                           ]
                       ]
                       [
                           Multi-Task/Domain/Objective/Auxiliary Learning  (\S\ref{subsec:multitask}),  fill=myyellow!60, text width=20.7em, text=black
                           [
                             {
                             Knowledge Transfer in MTL (\S\ref{subsubsec:kt_mtl}) \\
                             Knowledge Transfer in MOO (\S\ref{subsubsec:kt_moo}) \\
                             Knowledge Transfer in MDL (\S\ref{subsubsec:kt_mdl}) \\
                             Knowledge Transfer in ATL (\S\ref{subsubsec:kt_al})
                             }, 
                             color=hidden-draw, fill=myyellow!40, text width=13.8em, text=black
                           ]
                       ]
                       [
                           Out-of-Distribution/Domain Generalization (\S \ref{subsec:oodg_dg}),  fill=myyellow!60, text width=20.5em, text=black
                           [
                            {
                            Better OOD Generalization (\S \ref{subsubsec:oodg}) \\
                            Better DG Generalization (\S\ref{subsubsec:dg})
                            }, 
                            color=hidden-draw, fill=myyellow!40, text width=14em, text=black
                           ]
                       ]
                       [
                           Federated Learning (\S\ref{subsec:federated}),  fill=myyellow!60, text width=14em, text=black
                           [
                             Local Knowledge Aggregation (\S\ref{subsubsec:fl_local}), 
                             color=hidden-draw, fill=myyellow!40, text width=20.5em, text=black
                           ]
                       ]
                       [
                           Zero-shot/Few-Shot Learning (\S\ref{subsec:fewshot}), fill=myyellow!60, text width=14em, text=black
                           [
                            {
                            Zero-shot Knowledge Transfer (\S \ref{subsubsec:kt_zeroshow}) \\
                            Few-shot Knowledge Transfer (\S \ref{subsubsec:kt_fewshow})
                            }, 
                            color=hidden-draw, fill=myyellow!40, text width=20.5em, text=black
                           ]
                       ]
                       [
                           Adversarial Learning (\S\ref{subsec:adversarial}),  fill=myyellow!60, text width=14em, text=black
                           [
                            {
                            Model Attack (\S \ref{sssec:attack}) \\
                            Model Defense and Copyright Protection (\S \ref{sssec:defense})
                            },
                            color=hidden-draw, fill=myyellow!40, text width=20.5em, text=black
                           ]
                       ]
                ]
            ]   
        \end{forest}
    }
\caption{
The taxonomy of model merging in machine learning. 
This general framework covers advanced model merging methods (top part), theoretical and empirical analysis (middle part), and practical applications of model merging techniques to foundation models and more than 10 machine learning subfields (bottom part).
}
\label{fig:taxonomy}
\end{figure*}

\textbf{Second, which applications can benefit from model merging?}
We discuss in detail the various use cases of model merging in foundation models  (\S \ref{sec:application_lm}) and over ten subfields of machine learning (\S \ref{sec:application_ml}). As shown in Figure~\ref{fig:taxonomy} (lower part), model merging can be applied to a variety of foundation models, including large language models, multimodal large language models, and visual generative models. For example, model merging in large language models can help mitigate untruthfulness and toxicity output, accomplish knowledge unlearning, and speed up training. Moreover, model merging also arises in different machine learning subfields, such as continual learning, multi-task learning, few-shot learning, and other subfields, to solve a variety of challenges. For instance, in continual learning, model merging can mitigate catastrophic forgetting of old tasks. In multi-task/objective/domain learning, it facilitates knowledge transfer. Additionally, in adversarial learning, model merging can be employed for both attack and defense strategies.

\textbf{Third, what are the remaining challenges and future research opportunities for model merging?} 
Despite the advancements in merging methods and their well-developed applications, there are still numerous open challenges and future research directions in the field (\S \ref{sec:future_directions}). For example, as the number of tasks increases, the performance gap between existing methods and independent expert models becomes significantly larger. Additionally, current model merging methods incur enormous memory costs during merging and lack trust guarantees as well as in-depth theoretical analysis. Addressing these gaps will require substantial efforts from researchers to further advance the flourishing development of this field.

To summarize, the \textbf{main contributions} of this paper include the following three aspects:
\begin{itemize}
    \item \textit{Methodology Overview}:  We present a taxonomy of model merging methods that splits them into two stages and further classifies each stage by key techniques. We also review existing theoretical analyses and empirical evaluations.
     \item \textit{Application Overview}: We survey applications of model merging in foundation models and more than ten machine learning subfields, showing how it helps address existing challenges.
     \item \textit{Future Directions}: We discuss open challenges and future directions for model merging, including performance gaps, theoretical understanding, trustworthy guarantees, and cross-disciplinary applications.
\end{itemize}
The \textbf{structure} of this paper is as follows: \S\ref{sec:introduction} is an introduction, and \S\ref{sec:methods} offers a comprehensive discussion of advanced model merging methods from a technical perspective. \S\ref{sec:theory} validates representative model merging methods from both theoretical and experimental perspectives.
In \S\ref{sec:application_lm} and \S\ref{sec:application_ml}, we summarize the applications of model merging in various foundation models and different subfields within machine learning, respectively. 
The remaining challenges and future research directions are discussed in \S\ref{sec:future_directions}. Finally, \S\ref{sec:conclution} concludes the paper.

\section{Advanced Model Merging Methods}
\label{sec:methods}

In this section, we first introduce the notation and problem definition of model merging in \S\ref{subsec:notation}. We then elaborate on advanced model merging methods (Table~\ref{tab:method_summary} summarizes the primary purpose of each category of methods).
Existing model merging techniques can be roughly divided into the following two categories: (i) \textit{Before Merging Methods} in \S\ref{subsec:beforemerging}: it provides better prior knowledge for model merging. (ii) \textit{During Merging Methods} in \S\ref{subsec:duringmerging}: it resolves task conflict/interference by various strategies, and then performs parameter merging operations.

\begin{table*}[h]
\small
\caption{A summary of existing model merging methods.}
\vspace{-10pt}
\resizebox{\linewidth}{!}{ 
\begin{tabular}{c|c}
\toprule  
\textbf{Methods} &  \textbf{The Goal or Main Idea  of the Methods}    
\\ \midrule
\revised{Merger-Friendly Fine-tuning} (\S \ref{subsubsec:linearization})  & \revised{Design fine-tuning strategies that facilitate downstream model merging}
\\
Architecture Transformation (\S \ref{subsubsec:artchtecture})  & Transform multiple heterogeneous models into homogeneous models
\\
Weight Re-basin or Alignment (\S \ref{subsubsec:alignment})  & Repermutate multiple models into the same basin
\\ 
\midrule
Basic Merging Methods (\S \ref{subsubsec:basic})  & Simple weighted averaging or task-arithmetic based merging
\\ 
Weighted-based Merging Methods (\S \ref{subsubsec:weighted})  & Merge multiple models based on model/parameter importance weights
\\ 
Subspace-based Merging Methods (\S \ref{subsubsec:subspace})  & Merge multiple models by projecting them into a sparse or low-rank subspace
\\ 
\revised{Optimization-based Merging Methods (\S \ref{subsubsec:optimization})} & \revised{Construct data-free optimization objectives to mitigate model conflicts and interference}
\\ 
Routing-based Merging Methods (\S \ref{subsubsec:routing})  & Dynamically merge multiple models based on input during the inference phase
\\ 
Post-calibration-based Merging Methods (\S \ref{subsubsec:postmerging})  & Calibrating the merged model to be closer to the individual models reduces the knowledge loss
\\
\bottomrule
\end{tabular}
}
\label{tab:method_summary}
\vspace{-10pt}
\end{table*}

\subsection{Notation and Model Merging Problem Definition}
\label{subsec:notation}

Assume there are $T$ models ($\Phi_{\Theta^{(1)}},\ldots,\Phi_{\Theta^{(T)}}$) of the same architecture that need to be merged, and they train from scratch or fine-tune on the same pre-trained model $\Phi_{\Theta^{(0)}}$ respectively. The parameters (or weights) of the $t$-th model $\Phi_{\Theta^{(t)}}$ are represented as $\small \Theta^{(t)}=\{\Theta^{(t)}_l\}_{l=1}^L$, where $l$ denotes the $l$-th layer of the model, and $L$ is the total number of layers. 

In this survey, we focus on \textbf{parameter-wise} merging. In other words, the goal of model merging is to merge the parameters $\small \{\Theta^{(1)},\ldots,\Theta^{(T)}\}$, and finally obtain the new parameters $\small \Theta^{(merge)}=\texttt{merge}(\Theta^{(1)},\ldots,\Theta^{(T)})$.
One straightforward solution for merging models is {weighted averaging}~\cite{utans1996weight,shoemake1985animating}, defined as $\small \Theta^{(merge)} = \frac{1}{T} \sum_{t=1}^T \Theta^{(t)}$. However, the performance of this approach is often unacceptably poor or infeasible due to several possible factors: (i) The lack of suitable merging conditions, such as multiple models not being in the same basin or having inconsistent architectures. (ii) There are conflicts and interference among multiple models.  We illustrate how advanced methods address these issues in \S\ref{subsec:beforemerging} and \S\ref{subsec:duringmerging}, respectively.

\subsection{Pre-Merging Methods}
\label{subsec:beforemerging}
To better prepare models for merging, one class of work focuses on the fine-tuning and preprocessing of expert models, including \revised{linear fine-tuning to disentangle weights and sparse fine-tuning to reduce parameter conflicts during subsequent merging} (\S\ref{subsubsec:linearization}), transforming heterogeneous architectures into a common one (\S\ref{subsubsec:artchtecture}), and rebasining or aligning weights before merging (\S\ref{subsubsec:alignment}).

\subsubsection{\revised{Merger-Friendly Fine-Tuning}}
\label{subsubsec:linearization}
\begin{figure*}[t]
\centering  
\includegraphics[width=1\textwidth]{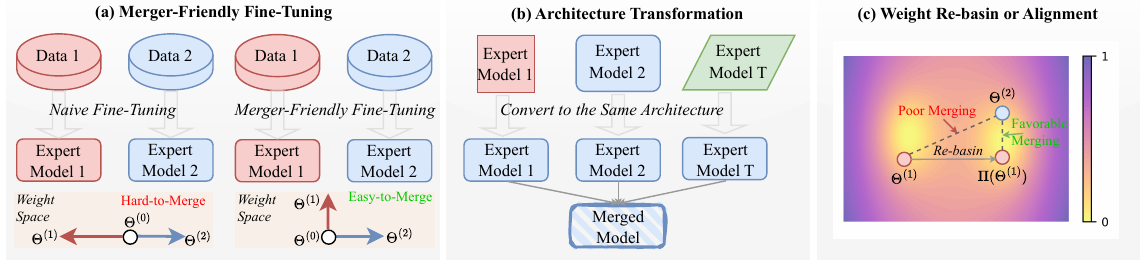}
\vspace{-20pt}
\caption{
(a) \revised{Illustration of merge-friendly fine-tuning: unlike standard fine-tuning, which yields models that are hard to merge, merge-friendly fine-tuning produces models that are easier to merge.}
(b) Illustration of an architectural transformation that converts multiple heterogeneous models into a homogeneous architecture, enabling direct parameter-level merging. 
(c) Illustration of weight/parameter alignment: permuting the neural network model $\small \Theta^{(1)}$ so that it aligns with $\small \Theta^{(2)}$.
}
\vspace{-10pt}
\label{fig:permuting_architectures}
\end{figure*}

\revised{
Merge-friendly fine-tuning designs training strategies that produce models more amenable to merging than standard fine-tuning, as shown in Figure~\ref{fig:permuting_architectures}(a), with minimal performance degradation after merging.
}
(i) \textbf{Linearization Fine-tuning.}
\citet{TangentSpace_NeurIPS2023} reveal that one necessary condition for effective model merging is `weight disentanglement'. This means that different directions of the weight space correspond to functional changes in disjoint regions of the input space. 
To achieve weight disentanglement, \citet{TangentSpace_NeurIPS2023} propose fine-tuning the linearized model along the tangent space~\cite{NTK_NeurIPS2018} of the pre-trained model during the fine-tuning stage, rather than in the original space of the nonlinear model. 
However, linearized fine-tuning with all parameters is more expensive than nonlinear fine-tuning. To accelerate this process, some works suggest linearizing only part of the layers. For example, \citet{LinearizationLoRA_ICLR2024} propose partially linearizing the Adapter modules and then merging Adapters. \citet{jin2024finetuninglinearlayerssimple} suggest linearly fine-tuning only the linear layers in the attention modules of the full model. Furthermore, TAFT~\cite{liu2024tangent} develops an efficient linearization method for the Transformer~\cite{transformer_2017} architectures, which directly derives closed-form linearized solutions for transformer networks.
\revised{
(ii) \textbf{Sharpness-Aware Fine-Tuning.} SAFT-Merge~\cite{leemitigating} adopts a sharpness-aware minimization to encourage a flatter task-specific loss landscape, thereby reducing the sensitivity of model parameters to merging-induced perturbations and better preserving performance after model merging.
}
\revised{
(iii) \textbf{Subspace Fine-Tuning.} Unlike the full-parameter fine-tuning strategies mentioned above, some approaches fine-tune only a subset of parameters instead of the whole set. Keeping most weights frozen during fine-tuning can reduce conflicts that arise in subsequent merging. For example, OSRM~\cite{zhang2025unraveling} fine-tunes the model via LoRA in an orthogonal subspace, thereby mitigating unintended interference across tasks. In addition, TaLoS~\cite{iuradaefficient} sparsely fine-tunes only a selected subset of parameters in the pretrained model to construct sparse task vectors with minimal interference.
}

\subsubsection{Architecture Transformation}
\label{subsubsec:artchtecture}

In some cases, models that need to be merged may have different architectures and cannot be merged directly. To solve this problem, some studies~\cite{avrahami2022ganCocktail,fusionllm,wan2024fusechat,nguyen2023cross} propose to perform architecture transformation before merging, that is, transform multiple models with different architectures into the same architecture as shown in Figure~\ref{fig:permuting_architectures}(b).
For example, GAN Cocktail~\cite{avrahami2022ganCocktail} attempts to merge multiple GAN models $\{\Theta^{(1)},\ldots,\Theta^{(T)}\}$ with different architectures. It transforms all GAN models $\small \Theta^{(t)}$ ($t \in \{1,2,\ldots,T\} \;\&\; t \!\neq \!k$) into a specified target model $\Theta^{(k)}$, that is, $\small \Theta^{(k)}$ is used as the initialization to learn the output of $\small \Theta^{(t)}$, while adding implicit regularizations to ensure that $\small \Theta^{(k)}$ does not forget knowledge of the task $k$. 
Similarly, FuseLLM~\cite{wanknowledge} and FusionChat~\cite{wan2024fusechat} propose to merge chat LLMs with diverse architectures and scales (e.g., NH2-Mixtral-8x7B~\cite{jiang2024mixtral}, NH2-Solar-10.7B~\cite{kim2023solar}, OpenChat-3.5-7B~\cite{wang2023openchat} in their practical applications). Specifically, FusionChat first uses knowledge distillation to transform all the architectures to match that of OpenChat-3.5-7B, and then performs the merge operation. 
Unlike the above distillation-based approach, CLAFusion~\cite{nguyen2023cross} adds layers/blocks (with weights set to the identity matrix) to the smaller model to align its architecture with that of the larger model.
In summary, merging models with different architectures requires first transforming the models into a common architecture.

\subsubsection{Weight Re-basin or Alignment}
\label{subsubsec:alignment}

The linear mode connectivity (LMC) property of deep neural networks demonstrates that there is a connected path between multiple local minima of deep neural networks along which the loss remains nearly constant~\cite{garipov2018loss,draxler2018essentially,tatro2020optimizing,lmc_iclr2022,ferbach2024proving,zhaounderstanding}. Numerous studies~\cite{frankle2020linear,nagarajan2019uniform,lmc_iclr2022,RotationSymmetry,xu2024training} have shown that two independent models, starting from the same pre-trained model and fine-tuned with different hyper-parameter configurations, typically satisfy LMC. Further, \citet{adilova2024layerwise} and \citet{zhou2023going} extended the study of LMC to the layer level. The LMC property implies that multiple local minima may be equivalent in the weight space, and different weight configurations of the same model may represent the same functionality. Inspired by this, many works proposed to permute the weights of one model (i.e., $\small \Theta^{(1)} \rightarrow \Pi(\Theta^{(1)})$) to align with the other model $\Theta^{(2)}$ when merging/interpolating two separate models, as illustrated in Figure~\ref{fig:permuting_architectures} (c). $\Pi(\cdot)$ denotes a permutation function, and researchers have dedicated efforts to studying effective and efficient permutation strategies for model alignment.
OTFusion~\cite{otfusion_neurips2020} and \citet{imfeld2024transformer} adopt optimal transport to soft-align neurons across models. NeuronAlignment~\cite{tatro2020optimizing} introduces an inexpensive heuristic algorithm to approximate the optimal neuron alignment. CCAMerge~\cite{horoi2024harmony} permutes by maximizing the correlation between linear combinations of neurons. Notably, Git re-basin~\cite{GitReBasin_ICLR2023} proposes three methods --activation matching, weight matching, and straight-through estimation-- to align (or permute) the weights of models trained on different tasks. Based on the Git re-basin, \citet{ReBasin_CVPR2023} further incorporate a Sinkhorn-based projection to improve these alignment methods. Unlike heuristic alignment strategies, Deep-Align~\cite{equivariant_ICML2024} proposes a learning-based approach to weight alignment, employing a novel learnable architecture that takes two sets of weights as input and outputs a permutation matrix for alignment. 
Despite the significant improvement of these alignment algorithms, \citet{jordan2022repair} argue that the success of these methods depends on the use of normalization layers (BatchNorm, LayerNorm, etc.) in the model; without these, the performance of the matching algorithms is greatly reduced. 
Additionally, \citet{c2m3} noted that previous pairwise permutations do not guarantee cycle consistency, making the alignment fragile. They further proposed to globally optimize the permutations of \textit{all} layers simultaneously at each step.
Overall, aligned models experience much less interference after merging compared to directly merging unaligned models.

\subsection{During Merging Methods}
\label{subsec:duringmerging}

In this section, we provide a detailed discussion on how to merge a set of well-trained expert models. The existing methods can be roughly divided into six categories: basic methods (\S\ref{subsubsec:basic}), weighted-based methods (\S\ref{subsubsec:weighted}), subspace-based methods (\S\ref{subsubsec:subspace}), optimization-based methods (\S\ref{subsubsec:optimization}), routing-based methods (\S\ref{subsubsec:routing}), and other merging methods (\S\ref{subsubsec:postmerging}).

\subsubsection{Basic Merging Methods}
\label{subsubsec:basic}
\begin{figure*}[t]
\centering  
\includegraphics[width=1\textwidth]{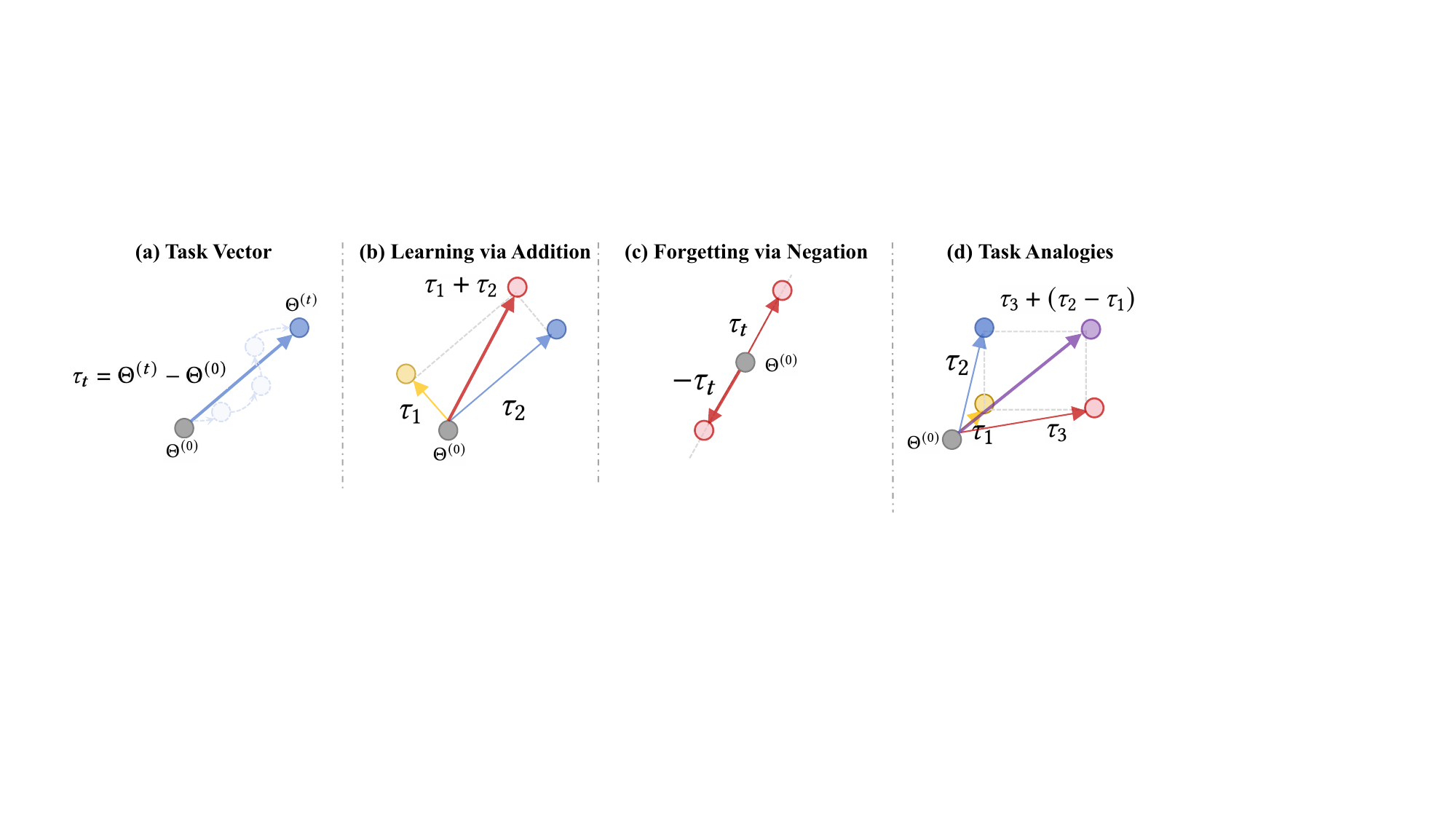}
\vspace{-15pt}
\caption{An illustration of Task Arithmetic~\cite{TaskArithmetic_ICLR2023}. (a) Definition of the ``task vector'', which is the difference between the fine-tuned model and the pre-trained model. (b) Multi-task learning is performed by merging multiple task vectors. (c) Knowledge forgetting is achieved by subtracting the task vector. (d) Analogical task vectors are used to implement task analogies.
}
\vspace{-10pt}
\label{fig:task_vector}
\end{figure*}

One of the most straightforward approaches to model merging is to directly weighted average the parameters of multiple models~\cite{utans1996weight,shoemake1985animating,Modelsoups_ICML2022}, i.e., $\small \Theta^{(merge)}=\frac{1}{T}  \sum_{t=1}^T \Theta^{(t)}$. 
Recently, Task Arithmetic~\cite{TaskArithmetic_ICLR2023} introduced the concept of ``task vector" (in Figure~\ref{fig:task_vector}(a)), which represents the model parameter $\Theta^{(t)}$ fine-tuned on task $t$ subtract the pre-trained model parameter $\Theta^{(0)}$, i.e., $\tau_t = \Theta^{(t)}-\Theta^{(0)}$. 
In other words, task vectors are thought to steer the behavior of a neural network meaningfully.
For example, multitask learning (MTL) can be accomplished by adding task vectors, forgetting can be achieved by subtracting task vectors, and task analogies can be performed using analogous task vectors. Specifically, when we want the pretrained model to perform MTL, we can add multiple task vectors $\{\tau_1,\ldots,\tau_T\}$ to the pretrained model, i.e., $\small \Theta^{(merge)}=\Theta^{(0)} + \lambda \cdot \sum_{t=1}^T \tau_t$ in Figure~\ref{fig:task_vector}(b), where $\lambda$ is a hyperparameter. Conversely, when we want the pretrained model to forget a function $t$, we can subtract the corresponding task vector from pretrained model as Figure~\ref{fig:task_vector}(c), i.e., $\small \Theta^{(merge)}=\Theta^{(0)} - \tau_t$. As shown in Figure~\ref{fig:task_vector}(d), we can also implement task analogies by task vector analogies, thus enabling zero-shot learning of new tasks. Similarly, PEMs~\cite{pem_neurIPS2023} combines Adapters with different capabilities by extending task arithmetic to parameter-efficient fine-tuning settings. However, the performance of basic merging methods is not satisfactory most of the time, especially when the tasks interfere with each other.

\subsubsection{Weighted-based Merging Methods}
\label{subsubsec:weighted}
\begin{figure*}[t]
\centering  
\includegraphics[width=1.0\textwidth]{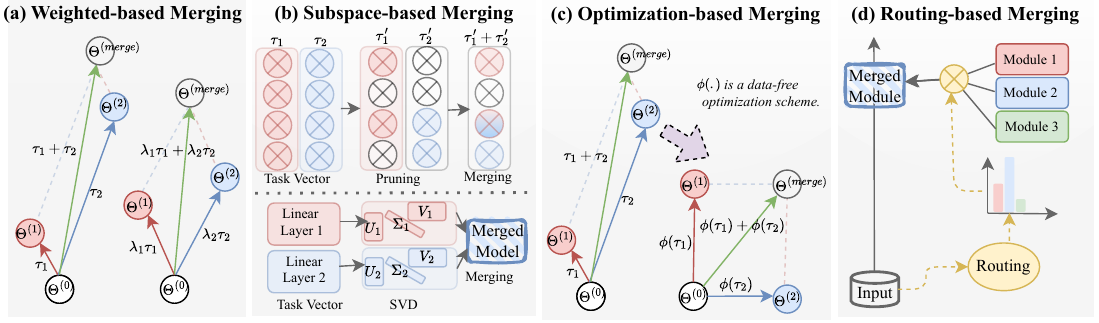}
\vspace{-15pt}
\caption{
An illustration of (a) weight-based model merging; (b) subspace-based merging, where empty entries denote zeros; (c) \revised{optimization-based merging}; and (d) routing-based merging that dynamically merges models based on the input.
}
\vspace{-15pt}
\label{fig:advanced_method}
\end{figure*}

Different models (or task vectors) represent different functions, and intuitively, different functions have varying degrees of importance. Therefore, weighted-based model merging methods design various clever rules to determine the merging coefficients~\cite{minut2025mergenetic,daileveraging}, as shown in Figure~\ref{fig:advanced_method}(a). 
For instance, when merging two models $\Theta^{(1)}$ and $\Theta^{(2)}$ (or task vectors $\tau_1$ and $\tau_2$), the goal of the weighted merging method is to find the optimal coefficients $\lambda_1^{*}$ and $\lambda_2^{*}$ so that the merged model $\small \Theta^{(merge)}=\lambda_1^{*} \Theta^{(1)}+\lambda_2^{*} \Theta^{(2)}$ (or $\small \Theta^{(merge)}= \Theta^{(0)} + \lambda_1^{*} \tau_1+\lambda_2^{*} \tau_2$) can retain the capabilities of the independent models as much as possible. A brute-force grid search to find the optimal merging coefficients is impractical when the number of models is large.

In order to determine the merging coefficient more effectively, Evolutionary-model-merge~\cite {akiba2024evolutionary} and Checkpoint Merging~\cite{liu2024checkpoint} efficient searches for the merging coefficients using evolutionary algorithms and Bayesian optimization, respectively. AdaMerging~\cite{AdaMerging_ICLR2024} uses gradient descent optimization to learn the merging coefficients by minimizing entropy as a surrogate loss in unlabeled test data. 
MetaGPT~\cite{zhou2024metagpt} casts the model merging problem as an MTL formalism, and it employs local linearization of the model and the orthogonality of the task vectors to derive the optimal merging coefficient $\lambda_t^{*}$ for each model $\tau_t$ as follows: $\small \lambda_t^{*}={\left\|\tau_t\right\|^2}/{\sum_{k=1}^T\left\|\tau_k\right\|^2}$. \revised{STF~\cite{qiu2025superpose} identifies task-specific features via singular value decomposition (SVD) and formulates the merging process as a linear system to solve for the optimal task-specific merging coefficients.}
SLERP~\cite{goddard2024ArceesMergeKit} performs spherical interpolation of the parameters of the two models. The interpolated coefficients of $\tau_1$ and $\tau_2$ are given by $\small \lambda_1^{*}=\frac{\sin \left(\left(1-\lambda \right) \cdot \rho \right)}{\sin (\rho)}$ and $\small \lambda_2^{*}\frac{\sin (\lambda  \cdot \rho)}{\sin (\rho)}$, respectively, where $\small  \rho  =\arccos \frac{\tau_1 \cdot \tau_2}{\left|\tau_1\right| \cdot\left|\tau_2\right|}$ denotes the angle between the two task vectors, and $\lambda$ represents the merging coefficient of the initial setting. However, SLERP can only merge two models.

The above weighting methods operate at the model (or task) level. It is well known that each layer and even each neuron in a deep neural network model plays a significantly different role, and some research has developed more fine-grained weighted merging strategies~\cite{wanglines}. For example, Layer-wise AdaMerging~\cite{AdaMerging_ICLR2024} and aTLAS~\cite{zhang2024knowledge} adaptively learn different sets of merging coefficients for each layer or module of the model, respectively. RegMean~\cite{RegMean_ICLR2023} indicates that closed-form solutions (relying on the data statistics provided by the training/validation set) exist for linear layers in model merging, while nonlinear layers can simply perform weight averaging. 
Other works utilize the Fisher information matrix~\cite{fisher1922mathematical} to assess the importance of parameters when merging~\cite{FisherMerging_NeurIPS2022,jhunjhunwala2024erasure,thennal2024fisher,tam2023merging}. Fisher-Merging~\cite{FisherMerging_NeurIPS2022} performs model merging based on the importance of the parameters in each independent model, that is, $\small \Theta^{(\text{merge})}=\sum_{t=1}^T F^{(t)} \Theta^{(t)} / \sum_{t=1}^T F^{(t)}$, where $F^{(t)}$ is the diagonal of the Fisher information matrix with respect to task $t$. 
\citet{daheim2024model} linked the inaccuracy of weighted average with gradient mismatch, and further proposed an uncertainty-based algorithm to reduce the matching error, ultimately merging the models based on a second-order Hessian estimation.

\subsubsection{Subspace-based Merging Methods}
\label{subsubsec:subspace}
Another class of advanced methods transforms models into sparse or low-rank subspaces for merging, thereby mitigating task interference, as shown in Figure~\ref{fig:advanced_method}(b).
(i) \textbf{\revised{Sparse Subspace Merging:}}
TIES-Merging~\cite{TiesMerging_NeurIPS2023} proposes to trim each individual model based on parameter magnitudes, retaining only the top 20\% of parameters with the highest magnitudes. It further suggests eliminating parameter sign conflicts to reduce interference, and finally merging sparse models using Task Arithmetic~\cite{TaskArithmetic_ICLR2023}. Similarly, DARE~\cite{DARE_Arxiv2023} also sparsifies by parameter magnitude, and highlights the importance of further performing rescaling on sparse models. In addition to removing the tail parameters with the smallest weight, the Model Breadcrumbs~\cite{davari2023modelbreadcrumbs} highlight the importance of removing the parameters (outliers) with the largest weights to further reduce noise in model merging and enhance generalization to hyperparameters. TALL-masks~\cite{wang2024localizingtall} creates a mask matrix specific to each task based on a predefined threshold related to independent models. Further, Model Tailor~\cite{zhu2024model} masks unimportant parameters based on the sensitivity of fine-tuned parameters to loss changes and the significance of changes compared to pre-trained parameters. APL~\cite{kong2024activated} proposes estimating parameter importance using causal interventions. \revised{PCB-Merging~\cite{du2024parameter} measures parameter importance within each task as well as the similarity of parameter importance across tasks, and discards parameters with low importance.} \revised{CALM~\cite{yancalm} integrates local information that is consistent with the global task consensus by aligning local binary masks with the global consensus and performing conflict-free merging.}
In contrast to the mask construction rules of the aforementioned heuristics, Concrete~\cite{tang2023concrete} frames mask construction and model merging as a learnable bi-level optimization problem. The outer-level optimizes the mask matrix, while the inner-level merges the model based on the mask matrix and optimizes it using the unlabeled test samples. 

\revised{
(ii) \textbf{Low-Rank Subspace Merging:} Low-rank subspace methods extract the most informative components via matrix factorization for merging, while discarding less important components that are more likely to cause conflicts.
TSV-M~\cite{gargiulo2025task} applies SVD to task vectors, concatenates the left and right singular vectors, and then orthogonalizes them to remove redundant bases. ISO-C~\cite{marczakno} further flattens the singular value spectrum of the task matrices.
HO-GSVD~\cite{skorobogat2025subspace} observes a rank-collapse phenomenon in the task-vector space during merging. To address this, it applies a subspace boosting procedure after generalized SVD to preserve the rank of the task vectors.
KnOTS~\cite{Knots} first concatenates all task vectors and performs SVD on the resulting high-dimensional matrix while enforcing a shared left singular matrix across tasks. It then merges the task-specific right singular matrices for different tasks. OPCM~\cite{tangmerging} projects the new task vector onto the orthogonal subspace of the previous task vectors to enable continual model merging.
}

\subsubsection{\revised{Optimization-based Merging Methods.}}
\label{subsubsec:optimization}
\revised{
Optimization-based methods exploit the intrinsic properties of task vectors to formulate an optimization objective, thereby enabling \textit{data-free} minimization of inter-task interference or information loss after merging, as shown in Figure~\ref{fig:advanced_method}(c).
For example, AWD~\cite{xiong2024multi} theoretically shows that interference between tasks can be minimized when task vectors are mutually orthogonal. Building on this insight, it proposes an adaptive weight disentanglement scheme to optimize the original task vectors, making them more orthogonal. DOGE~\cite{weimodeling} formulates model merging as a constrained optimization problem and interprets task vectors as accumulated gradients. It then solves this problem via an adaptive projected gradient descent algorithm, enabling data-free optimization of task vectors to alleviate conflicts. WUDI-Merging~\cite{wudimerging2025} shows that the task vectors of a linear layer span an approximate linear subspace of its corresponding inputs, and leverages this property to construct a data-free loss function that minimizes interference among task vectors. DOP~\cite{DOP_NeurIPS2025} combines data-space approximation with orthogonal projection, and employs a data-free loss function to balance stability and plasticity, thereby enabling continual model merging.
}

\subsubsection{Routing-based Merging Methods}
\label{subsubsec:routing}

The basic, weighted, subspace or optimization-based merging methods discussed in \S\ref{subsubsec:basic}--\S\ref{subsubsec:optimization} are \textit{static} merging methods. This means that the merged model remains the same for all samples or tasks. Given that there are differences between input samples/tasks, the model's ability may vary when processing different samples/tasks. As shown in Figure~\ref{fig:advanced_method} (d), some works propose to \textit{dynamically} merge models (or subsets of layers) based on the samples/tasks~\cite{li2023merge,muqeeth2024soft,tang2024moemerging,lu2024twin,kang2024self,ye2025dynamic} during the inference phase.
EMR-Merging~\cite{emrmerging_arxiv2024} proposes maintaining a shared model among multiple tasks alongside a sparse task-specific model. During inference, the sample-specific sparse task model is merged with the shared dense model.
In addition, for a given input, SMEAR~\cite{muqeeth2024soft} first computes a weighted average of the parameters of each expert by using the distribution of router inputs to the expert modules. The advantage of this approach is that it has a similar computational cost to that of a single expert. Twin-Merging~\cite{lu2024twin} also adaptively combines task-shared and task-private knowledge based on routing during the inference phase.
Similarly, WEMoE~\cite{tang2024moemerging} proposes a dynamic merging Transformer architecture. It uses a standard weighted average to merge all modules except the linear layer. The linear layer is dynamically weighted and merged according to the routing network during inference. PWE MoE~\cite{tang2024towards} further extends WEMoE to a multi-objective optimization setting and uses the preference vector as input for routing. \revised{MoW-Merging~\cite{ye2025dynamic} employs a gating network to adaptively generate merging coefficients conditioned on the input sample, thereby enabling sample-level dynamic merging and automatic classifier selection.}

\subsubsection{Post-calibration-based Methods.}
\label{subsubsec:postmerging}
\citet{surgery_icml2024} introduce a post-merging method to calibrate merged models. They observed that merged models suffer from representation bias, meaning the representations extracted by the independent and merged models are very different, leading to performance degradation in the merged model. To alleviate this problem, they propose a module called `representation surgery' to calibrate the representation bias. 
\revised{
Building on this, SurgeryV2~\cite{yang2024surgeryv2} assumes that representation bias arises at every layer of the network and therefore applies the representation surgery strategy layer-wise.
ProbSurgery~\cite{wei2025representation} argues that the deterministic modeling in Surgery is suboptimal and introduces a probabilistic formulation to better eliminate representation bias. Similarly, LOT-Merging~\cite{sun2025towards} adopts a layer-wise scheme that explicitly minimizes the feature drift between task-specific expert models and the merged model.
}

\subsection{\revised{Summary and Discussion}}
\label{subsec:summary}

\revised{
Table \ref{tab:method_sumamry} summarizes and compares several representative model merging methods. We make the following key observations:
(1) Compared with maintaining a set of independent expert models, model merging can significantly reduce the total number of parameters that need to be stored and managed, especially when the number of tasks $N$ is large.
(2) Training-free methods and training-based methods each have their own advantages. Training-free approaches are simpler and easier to deploy across different application scenarios, whereas training-based methods require additional computational resources and come with higher implementation complexity.
(3) Most existing works do not require extra data at the merging stage, but the lack of data-driven merging may limit performance. In contrast, data-dependent approaches rely on labeled training or validation data, or on unlabeled test data, to guide the merging process.
(4) The vast majority of model merging methods adopt a static scheme, which introduces no additional overhead at inference time. Dynamic merging, however, must first compute sample- or task-dependent merged parameters before prediction, incurring extra latency. Moreover, dynamic approaches often require storing additional expert parameters, thereby increasing memory usage.
(5) The weighting granularity in model merging ranges from coarse to fine, including global, task-wise, layer-wise, and parameter-wise weighting. Finer-grained weighting schemes generally deliver better performance but typically rely on extra data to estimate appropriate merging coefficients. By contrast, many optimization-based and low-rank subspace methods are designed to work in a largely data-free manner.

Overall, there is no single “perfect” merging method that fits all scenarios; the most appropriate approach should be chosen based on performance requirements, implementation complexity, and the availability of data and computational resources.
}

\begin{table}[t]
\centering
\caption{
\revised{
A brief overview of representative model merging methods. 
\textit{Category}: The methodological category each approach belongs to.
\textit{Parameters}: The number of parameters in the merged model, where $1\times$ denotes the parameter size of a single-task model and  $N$ denotes the total number of tasks.
\textit{Tuning-Free}: Whether the merging process is tuning-free.
\textit{Required Data}: What additional data (if any) are required for merging.
\textit{Pattern}: Whether the merging pattern is static or dynamic.
\textit{Weighted Level}: The granularity at which multiple models are weighted during merging.
Here, \cmark=yes and \xmark=no.
}
}
\vspace{-10pt}
\resizebox{\linewidth}{!}{ 
\begin{tabular}{l|ccccccc}
\toprule
\textbf{Method} & \textbf{Category} & \textbf{Parameters} & \textbf{Tuning-Free} & \textbf{Required  Data} & \textbf{Pattern} & \textbf{Weighted Level}  \\
\midrule
\rowcolor{myultralightgray}
Traditional MTL &-  & $=1\times$ & \xmark & Labeled Training Dataset  & Static & -  \\
\rowcolor{myultralightgray}
Individual Models &- & $=N\times$ & \xmark & Labeled Training Dataset & Static & - \\
\midrule
\rowcolor{myultralightblue}
Weighted Average~\cite{Modelsoups_ICML2022} &Basic  & $=1\times$ & \cmark & \xmark & Static & Global \\
\rowcolor{myultralightblue}
Task Arithmetic~\cite{TaskArithmetic_ICLR2023}&Basic  & $=1\times$ & \cmark & \xmark & Static & Global \\
\midrule
\rowcolor{myultralightpurple}
SLERP~\cite{goddard2024ArceesMergeKit}&Weighted & $=1\times$ & \cmark & \xmark & Static & Task \\
\rowcolor{myultralightpurple}
Evolutionary-model-merge~\cite{akiba2024evolutionary}&Weighted &$=1\times$  &\cmark  & Labeled Validation Dataset & Static & Layer \\
\rowcolor{myultralightpurple}
AdaMerging~\cite{AdaMerging_ICLR2024}&Weighted & $=1\times$ & \xmark & Unlabeled Test Dataset & Static & Task/Layer \\
\rowcolor{myultralightpurple}
MetaGPT~\cite{zhou2024metagpt}&Weighted &$=1\times$  &\cmark  &\xmark & Static & Task \\
\rowcolor{myultralightpurple}
Fisher-Merging~\cite{FisherMerging_NeurIPS2022}&Weighted &$=1\times$  & \xmark   & Labeled Validation Dataset & Static & Parameter \\
\rowcolor{myultralightpurple}
RegMean~\cite{RegMean_ICLR2023}&Weighted &$=1\times$  & \xmark   & Labeled Validation Dataset & Static & Parameter \\
\midrule
\rowcolor{myultralightgreen}
Ties-Merging~\cite{TiesMerging_NeurIPS2023}&Sparse Subspace & $=1\times$ & \cmark &  \xmark & Static & Global \\
\rowcolor{myultralightgreen}
DARE~\cite{DARE_Arxiv2023} &Sparse Subspace& $=1\times$ & \cmark &  \xmark & Static & Global \\
\rowcolor{myultralightgreen}
Model Breadcrumbs~\cite{davari2023modelbreadcrumbs}&Sparse Subspace &$=1\times$  & \cmark &  \xmark &Static & Global \\
\rowcolor{myultralightgreen}
Model Tailor~\cite{zhu2024model}&Sparse Subspace &$=1\times$  &\xmark  & Labeled Training Dataset & Static & Global \\
\rowcolor{myultralightgreen}
Localize-and-Stitch~\cite{he2024localize} &Sparse Subspace &$=1\times$ & \cmark &  Labeled Validation Dataset & Static & Global \\
\rowcolor{myultralightgreen}
Consensus Merging~\cite{wang2024localizingtall} &Sparse Subspace &$=1\times$ & \cmark &  \xmark & Static & Task \\
\rowcolor{myultralightgreen}
DELLA-Merging~\cite{deep2024della} &Sparse Subspace &$=1\times$ & \xmark &  \xmark & Static & Global \\
\rowcolor{myultralightgreen}
PCB-Merging~\cite{du2024parameter}&Sparse Subspace &$=1\times$ & \cmark &  \xmark & Static & Task \\
\rowcolor{myultralightgreen}
KnOTS~\cite{Knots} &Low-Rank Subspace&$=1\times$  & \cmark &  \xmark & Static & Global \\
\rowcolor{myultralightgreen}
HO-GSVD~\cite{skorobogat2025subspace}&Low-Rank Subspace&$=1\times$  &  \cmark&  \xmark & Static & Global \\
\rowcolor{myultralightgreen}
TSV-M~\cite{gargiulo2025task} &Low-Rank Subspace&$=1\times$  &  \cmark&  \xmark & Static & Global \\
\rowcolor{myultralightgreen}
ISO-Merging~\cite{marczakno} &Low-Rank Subspace& $=1\times$ &\cmark  &  \xmark & Static & Global \\
\rowcolor{myultralightgreen}
OPCM~\cite{tangmerging} &Low-Rank Subspace& $=1\times$ &\cmark  &  \xmark & Static & Task \\
\midrule
\rowcolor{myultralightyellow}
AWD~\cite{xiong2024multi} &Optimization & $=1\times$ &  \xmark & \xmark & Static & Task \\
\rowcolor{myultralightyellow}
DOGE~\cite{weimodeling} &Optimization & $=1\times$ &  \xmark & \xmark & Static & Task \\
\rowcolor{myultralightyellow}
WUDI-Merging~\cite{wudimerging2025} &Optimization& $=1\times$ & \xmark &\xmark  & Static & Global \\
\rowcolor{myultralightyellow}
DOP~\cite{DOP_NeurIPS2025} &Optimization&$=1\times$  & \xmark & \xmark & Static & Task \\
\midrule
\rowcolor{myultralightbrown}
Surgery~\cite{surgery_icml2024} &Calibration & $>1\times$ & \xmark & Unlabeled Test Dataset & Static & Global \\
\rowcolor{myultralightbrown}
ProbSurgery~\cite{wei2025representation}&Calibration & $>1\times$ & \xmark & Unlabeled Test Dataset & Static & Global \\
\midrule
\rowcolor{myultralightpink}
EMR Merging~\cite{emrmerging_arxiv2024}&Routing & $> 1\times$ &\cmark   & \xmark & Dynamic & Task \\
\rowcolor{myultralightpink}
Twin Merging~\cite{lu2024twin} &Routing& $> 1\times$ &  \xmark & Labeled Validation Dataset & Dynamic & Sample \\
\rowcolor{myultralightpink}
WEMoE~\cite{tang2024moemerging}&Routing & $\gg 1\times$ &  \xmark & Unlabeled Test Dataset & Dynamic & Sample \\
\bottomrule
\end{tabular}
}
\label{tab:method_sumamry}
\vspace{-15pt}
\end{table}

\section{Theoretical and Experimental Analysis of Model Merging}
\label{sec:theory}

\subsection{Theoretical Analysis}
\label{subsec:theory}

In addition to designing various advanced methods in \S\ref{subsec:beforemerging} and \S\ref{subsec:duringmerging}, the theoretical and effectiveness analysis of model merging is also crucial. Currently, there is limited work on the theoretical analysis of model merging. Based on the source of the models to be merged, the existing theoretical analysis can be roughly divided into three categories.

(i) \textbf{Single-Trajectory Model Averaging.} Some analyses target model merging on the single-trajectory training, usually referring to stochastic weighted average (SWA) or exponential moving average (EMA). For example, \citet{jain2018parallelizing} theoretically proved that the excess risk of the EMA is an upper bound of a bias term and a variance term in the context of least squares regression. The bias term depends on the initialization state of the parameters and decreases exponentially with the number of iterations once the model starts averaging. The variance term depends on the noise covariance inherent in the data, which decays at a faster rate when model averaging is used~\cite{arpit2022ensemble}. Similarly, \citet{rame2022diverse} applies bias-variance decomposition to the domain generalization setting to explain why model averaging improves out-of-distribution performance. In addition, \citet{hardt2016train} provide a stability bound for SWA under convex assumptions, while \citet{wanggeneralization} further establish generalization bounds analysis in both convex and nonconvex cases.

(ii) \textbf{Same-Task Merging.} Some studies explain the merging of multiple models with different hyperparameter fine-tuning for the same dataset in terms of connectivity and flatness of the loss landscape~\cite{kuditipudi2019explaining,simsek2021geometry,wang2022plateau,benton2021loss}. Specifically, some works apply the theory of linear mode connectivity (LMC)~\cite{garipov2018loss,draxler2018essentially,tatro2020optimizing} of neural networks to explain model merging. 
\revised{
Here, we first give a formal definition of LMC~\cite{nagarajan2019uniform}. Given a dataset $\small \mathcal{D}$ and two trained models $\small \Theta^{(i)}$ and $\Theta^{(j)}$ such that $\small \mathcal{L}(\Theta^{(i)}) \approx \mathcal{L}(\Theta^{(j)})$ on $\mathcal{D}$, if there exists $\lambda \in [0,1]$ satisfying the following condition, we say that $\small \Theta^{(i)}$ and $\small \Theta^{(j)}$ are linearly connected in the loss landscape:
$
\small
\mathcal{L}\left(\lambda \Theta^{(i)} +(1-\lambda) \Theta^{(j)} \right) \approx \mathcal{L}\left(\Theta^{(i)}\right) \approx \mathcal{L}\left(\Theta^{(j)}\right) .
$
LMC indicates that the loss minima of neural networks are not isolated points in weight space; from the perspective of LMC, these minima can be regarded as equivalent.
}
Recent studies~\cite{frankle2020linear,nagarajan2019uniform,lmc_iclr2022,CTL2024} have shown that two independent models, starting from the same pre-trained model and fine-tuned with different configurations, usually satisfy LMC. 
\revised{Therefore, performing weight alignment following LMC can transform multiple models to be merged into a parameter basin with higher merging feasibility, thereby providing a robust validity guarantee for model merging~\cite{GitReBasin_ICLR2023,jordan2022repair}.}
On the other hand, other studies explain model merging from the perspective of a flatter loss landscape~\cite{li2018visualizing}, arguing that merging multiple weights fine-tuned under different optimization configurations with the same data usually converges to a flat local minimum~\cite{foret2021sharpness}, thus revealing why model merging has better generalization~\cite{smith2017investigation,izmailov2018averaging,SWAP_ICLR2020,zhang2020swa,SWAD_NeurIPS2021}. 

(iii) \textbf{Cross-Task Merging.} An analysis by \citet{TangentSpace_NeurIPS2023} is based on multiple models fine-tuned on different datasets, identifying weight disentanglement as a necessary precondition for effective task vector based merging. 
\revised{
More specifically, weight disentanglement is formally defined as follows. 
Consider a parametric function $\small f : \mathcal{X} \times \Theta \rightarrow \mathcal{Y}$. We say that $f$ is weight-disentangled if, for a set of task vectors $\small \{\tau_t\}_{t=1}^T$ and the corresponding datasets $\small \{\mathcal{D}_t\}_{t=1}^T$, the following holds:
$
\small
f\left({x} ; \Theta^{(0)}+\sum_{t=1}^T \lambda_t {\tau}_t\right)=\sum_{t=1}^T g_t\left({x} ; \lambda_t {\tau}_t\right)+g_0({x}), \; t=1,2,\ldots,T,
$
where $\small g_t({x}; \lambda_t {\tau}_t) = 0$ if the sample $x$ does not belong to task $t$, i.e., $\small x \notin \mathcal{D}_t$, and $\small \small g_0({x}) = 0$ if $x$ does belong to task $t$. Intuitively, each task vector ${\tau}_t$ contributes only to its own task-specific component $g_t$, while $g_0$ captures the task-agnostic part.
}
\citet{TangentSpace_NeurIPS2023} provide theoretical and empirical analyses of the neural tangent kernel (NTK) and establish a compelling link between the task arithmetic~\cite{TaskArithmetic_ICLR2023} and the spectral properties of NTK. \revised{From the NTK formulation, task vectors can be viewed as directions in the NTK feature space; when they are "approximately orthogonal" under the corresponding metric, they do not hinder each other during merging. 
}
\revised{In addition, \citet{zhou2025task} establish a connection between task vectors and task gradients. Under standard full-batch gradient descent, the task vector obtained after one epoch of fine-tuning is exactly equal to the negative gradient of the loss scaled by the learning rate, i.e., $\small \tau_t = \eta \nabla \mathcal{L}_t(\Theta^{(0)})$. For multi-epoch fine-tuning, this relationship becomes approximate rather than exact.
\citet{wang2025more} use Approximate Kinematics Theory to derive an upper bound on the number of models that can be effectively merged. Once this threshold is exceeded, further merging no longer leads to performance improvements.
}

\revised{
\textbf{Summary and Discussion.} 
Overall, the LMC perspective provides a direct explanation of why linear interpolation becomes reasonable once model weights are properly aligned. However, most existing analyses are conducted under the assumption of shared tasks or shared initialization, which limits their ability to account for the effectiveness of merging across heterogeneous tasks and modalities. In contrast, the connection between flat minima in the loss landscape and model generalization offers a plausible explanation for why weight averaging can improve out-of-distribution (OOD) performance, although a clear theoretical link between flatness and mergeability is still lacking. Finally, NTK-based theory explains how the similarity or approximate orthogonality between task vectors can mitigate conflicts during merging, but most of these results rely on wide-network limits and linearization assumptions, and their validity for large-scale LLM settings remains to be empirically verified.
Taken together, these theoretical lines emphasize complementary yet distinct perspectives, and a unified theoretical framework for model merging has not yet emerged—an important direction for future research.
}

\subsection{\revised{Empirical Comparisons}}
\label{subsec:EmpiricalComparisons}

\subsubsection{\revised{Benchmarks}}

\revised{
To enable a fair comparison of different model merging methods, a number of benchmark-style works have been proposed. In Table \ref{tab:benchmarkcomparison}, we present a comparison of existing model merging benchmarks. Specifically, MergeKit~\cite{arceemergekit2024} is one of the earliest toolkits for merging model checkpoints. It combines multiple LLMs to improve performance and generality, and most of the merging strategies it includes are classic, easy-to-implement methods. FusionBench~\cite{tangFusionBenchComprehensiveBenchmark2024} provides a model merging benchmark on visual classification tasks, but offers only limited support for LLM-based merging. Realistic Evaluation~\citep{tam2024realistic} instead focuses on a benchmark for merging models on visual generation tasks.
In recent years, more benchmarks have shifted their attention to LLMs. MergeBench~\cite{he2025mergebench} and \citet{hitit2025systematic} merge multiple domain-specialized LLMs, while Model-GLUE~\citep{ModelGLUE} defines a selection and aggregation protocol over a heterogeneous model zoo with different architectures and initializations. OptMerge~\cite{he2025mergebench}  further targets the merging of multimodal large language models.
In addition, several works analyze model merging from different perspectives. Merging at Scale~\citep{yadav2024matters} and Merging Scaling Law~\cite{wang2025model} investigate how the number of merged models affects performance, but mainly focus on classical merging methods. Mergenetic~\citep{minut2025mergenetic} instead studies the impact of evolutionary algorithms on merging performance.
}

\begin{table}[t]
\centering
\caption{\revised{Comparison of existing model merging benchmarks. Columns indicate whether each benchmark includes vision models (image classification/generation), language models (standard NLP tasks), large language models (complex text generation), multimodal large language models (text + vision/audio), and whether the benchmark is open-sourced. Here, \cmark=yes and \xmark=no.
}} 
\vspace{-10pt}
\label{tab:benchmarkcomparison}
\resizebox{\linewidth}{!}{ 
\revised{
\begin{tabular}{lccccccc}
\toprule
Evaluation & Vision Model & Text Model & Large Language Model & Multimodal Large Language Model & Open-Source \\
\midrule
MergeKit~\cite{arceemergekit2024}&\xmark   &\xmark   &\cmark  &\xmark  &\cmark  \\
FusionBench~\citep{tangFusionBenchComprehensiveBenchmark2024} &\cmark   &\cmark   &\cmark  &\xmark  &\cmark  \\
Realistic Evaluation~\citep{tam2024realistic} &\cmark   &\cmark   &\xmark  &\xmark  &\cmark  \\
Merging at Scale~\citep{yadav2024matters}  &\xmark   &\xmark   &\cmark  &\xmark  &\xmark  \\
Model-GLUE~\citep{ModelGLUE} &\xmark   &\xmark   &\cmark  &\xmark  &\cmark  \\
MergeBench~\cite{he2025mergebench} &\xmark   &\xmark   &\cmark  &\xmark  &\cmark  \\
Mergenetic~\citep{minut2025mergenetic}  &\xmark   &\xmark   &\cmark  &\xmark  &\cmark  \\
Merging Scaling Law~\cite{wang2025model} &\xmark   &\xmark   &\cmark  &\xmark  &\xmark  \\
Systematic Study of Model Merging~\cite{hitit2025systematic}&\xmark   &\xmark   &\cmark  &\xmark  & \xmark  \\
OptMerge~\cite{wei2025unifying} &\xmark   &\xmark   &\xmark  &\cmark  &\cmark  \\
\bottomrule
\end{tabular}
}
}
\vspace{-10pt}
\end{table}

\begin{figure*}[t]
\centering  

\begin{subfigure}{0.47\textwidth}
    \centering
    \includegraphics[width=\linewidth]{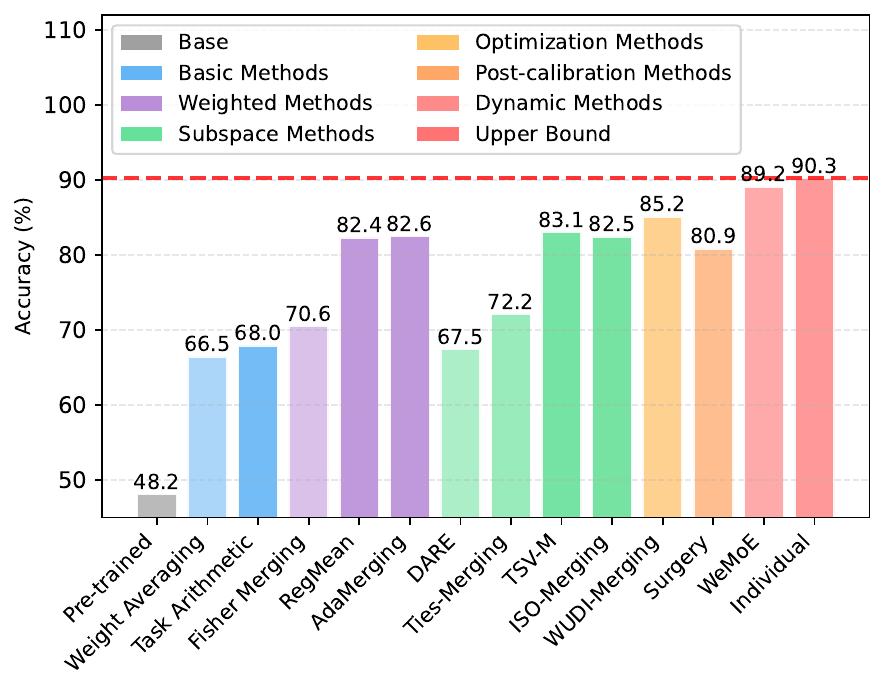}
    \vspace{-25pt}
    \caption{ViT-B/32}
    \label{fig:vitb32_acc}
\end{subfigure}
\hfill
\begin{subfigure}{0.47\textwidth}
    \centering
    \includegraphics[width=\linewidth]{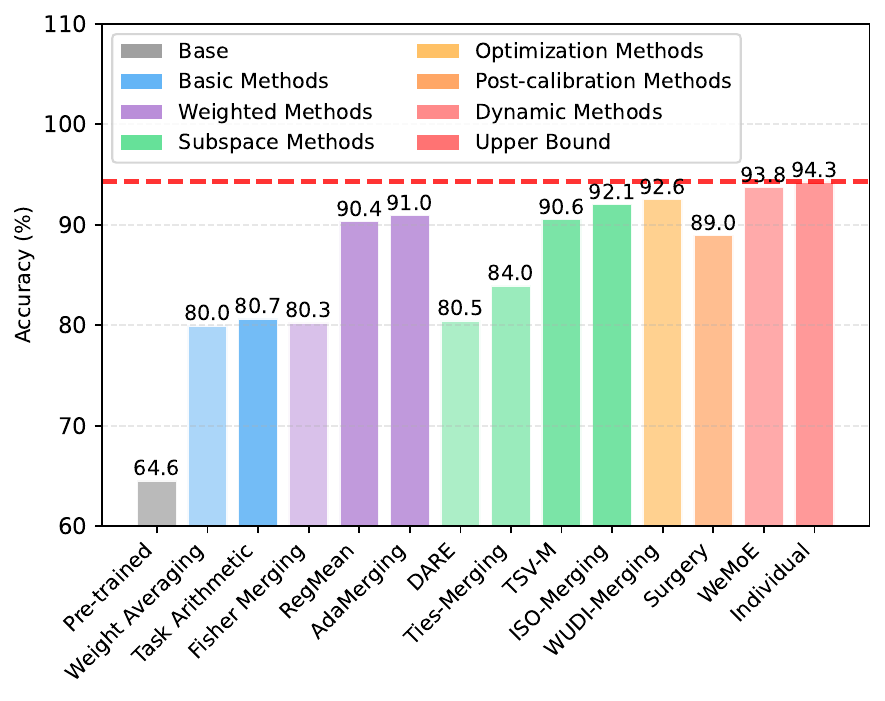}
    \vspace{-25pt}
    \caption{ViT-L/14}
    \label{fig:vitl14_acc}
\end{subfigure}

\begin{subfigure}{0.253\textwidth}
    \centering
    \includegraphics[width=\linewidth]{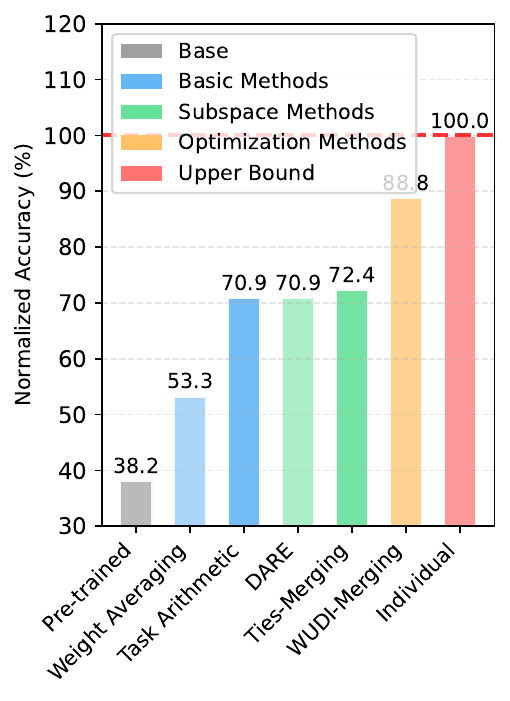}
    \vspace{-25pt}
    \caption{RoBERTa-Large}
    \label{fig:roberta_large_acc}
\end{subfigure}
\hfill
\begin{subfigure}{0.29\textwidth}
    \centering
    \includegraphics[width=\linewidth]{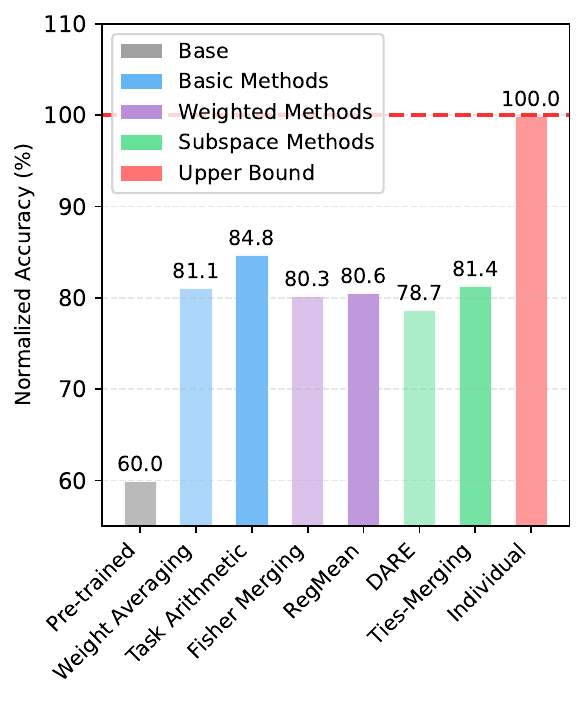}
    \vspace{-25pt}
    \caption{Llama-3.1}
    \label{fig:llama31_8b_acc}
\end{subfigure}
\hfill
\begin{subfigure}{0.403\textwidth}
    \centering
    \includegraphics[width=\linewidth]{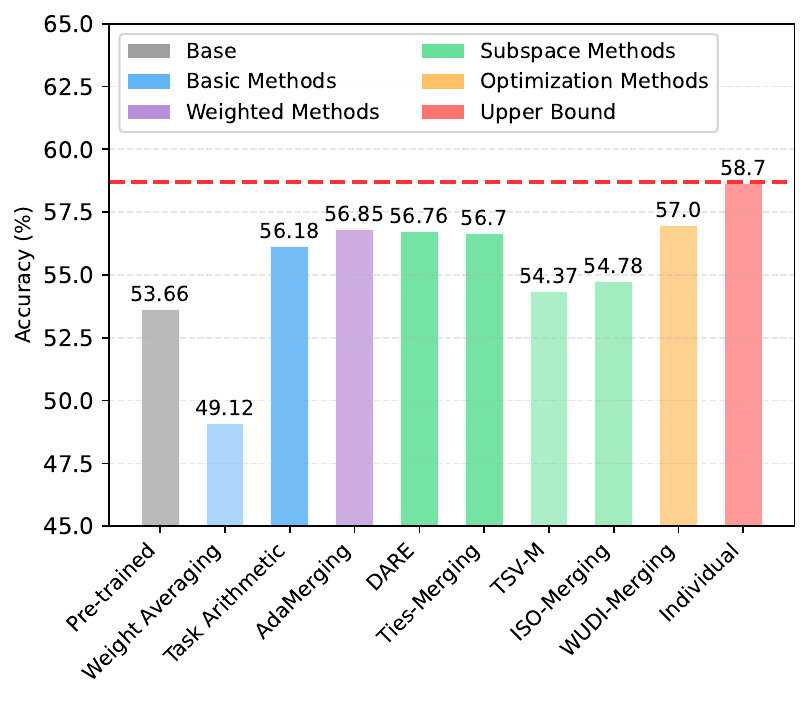}
    \vspace{-25pt}
    \caption{InternVL2.5}
    \label{fig:internvl25_acc}
\end{subfigure}

\vspace{-10pt}
\caption{\revised{Accuracy of Various Model Merging Methods on Five Backbones: 
(a) ViT-B/32, (b) ViT-L/14, (c) RoBERTa-Large, (d) Llama-3.1-8B, (e) InternVL-2.5. 
Specifically, 8 visual classification tasks are merged on ViT-B/32 and ViT-L/14, 8 NLP tasks on RoBERTa-Large, 5 text generation tasks on Llama-3.1, and 5 multimodal tasks on InternVL-2.5. 
}}
\vspace{-15pt}
\label{fig:acc_performance}
\end{figure*}

\subsubsection{\revised{Comparison of Results}}

\revised{
To gain a more intuitive understanding of the performance differences between various model merging methods, we present the performance of common merging methods across different domains.

\textbf{Visual Model Merging.} Two architectures, ViT-B/32 and ViT-L/14, were selected, and their performance when merging 8 tasks was evaluated following the Fusionbench~\citep{tangFusionBenchComprehensiveBenchmark2024} protocol. As shown in Figures  \ref{fig:vitb32_acc} and \ref{fig:vitl14_acc}, several key observations emerge: (1) Pre-trained models exhibit poor performance, as they lack task-specific capabilities tailored to downstream tasks. (2) Independent models (trained separately for each task) achieve the best performance due to the absence of cross-task interference. (3) Simple merging methods (e.g., Weight Averaging) result in significant performance degradation compared to independent models. (4) Weighted merging methods (e.g., RegMean, AdaMerging) and subspace merging methods (e.g., TSV-M, ISO-Merging) substantially improve the performance of merged models. (5) Optimization-based WUDI-Merging and post-hoc alignment method Surgery also demonstrates excellent performance. (6) The dynamic merging method WEMoE achieves performance closest to the upper bound. (7) Models with larger parameter scales are generally easier to merge (i.e.,  ViT-L/14 v.s. ViT-B/32).

\textbf{Language Model Merging.} Following \citet{wudimerging2025}, RoBERTa-Large models trained on eight GLUE tasks are merged, with the results reported in Figure \ref{fig:roberta_large_acc}. (1) The pretrained model and individually fine-tuned models still serve as the lower and upper performance bounds, respectively. (2) Subspace-based merging methods do not yield a significant performance change compared with Task Arithmetic. (3) In contrast, the optimization-based WUDI-Merging method achieves a clear performance improvement.

\textbf{Large Language Model Merging.} Following MergeBench~\cite{he2025mergebench}, five domain-specialized LLMs are merged, with the results reported in Figure \ref{fig:llama31_8b_acc}. (1) All merging methods lie between the pretrained model and the individually fine-tuned models, indicating that merging is beneficial but still incurs a certain degree of knowledge loss. (2) More sophisticated merging strategies, such as subspace-based or weighted merging, can even underperform the basic Task Arithmetic baseline. This may be due to the removal of neurons from the task vectors, which discards important information, or to the limited number of samples, which makes it difficult to reliably estimate parameter importance. In particular, DARE prunes neurons at random, leading to a high probability of removing important neurons and thus degrading performance.

\textbf{Multimodal Large Language Model Merging.} Following OptMerge~\cite{he2025mergebench}, multimodal models trained on five tasks are merged, and the results are reported in Table \ref{fig:internvl25_acc}. (1) Naive parameter averaging leads to severe performance degradation, even performing worse than the pretrained model. (2) Weighted merging achieves a substantial improvement over simple averaging, and sparse subspace methods such as DARE and TIES-Merging also perform well, whereas low-rank subspace approaches like TSV-M and ISO-Merging offer only moderate gains. (3) The optimization-based WUDI-Merging still achieves the best performance among all merging methods. (4) Nevertheless, all model merging approaches remain noticeably behind the individually fine-tuned models.

\textbf{Summary and Discussion.} These results suggest that simple model merging strategies typically deliver only moderate performance, whereas more advanced approaches—such as subspace-based or weighted merging—can further alleviate conflicts during merging and thus yield better accuracy. Optimization-based methods and dynamic merging schemes usually achieve the strongest results. However, all merging methods still fall short of the upper bound given by individual models.
}

\section{Application of Model Merging in Foundation Models}
\label{sec:application_lm}

The emergence of foundation models, including large language models (LLMs), multimodal large language models (MLLMs), and image generative models, is a significant indicator of technological progress in the field of artificial intelligence in recent years. However, despite their advancements, these large models still face several challenges, such as generating harmful content in LLMs, MLLMs struggling with fusing information from different modalities, and the difficulty of producing mixed-style images in image generation models. Recent studies suggest that model merging techniques offer a promising solution to these inherent challenges in foundational models. Table~\ref{tab:application_summary_lm} briefly summarizes the application of model merging in foundational models.

\begin{table*}[h]
\small
\caption{A summary of the application of model merging techniques in foundation models.}
\vspace{-10pt}
\resizebox{\linewidth}{!}{ 
\begin{tabular}{c|c}
\toprule  
\textbf{Scenarios} &  \textbf{The Main Purpose of Model Merging}    
\\ \midrule
Large Language Models (\S\ref{subsec:llms}) &  Enhancing the domain-specific capabilities of pre-trained LLMs  or editing old knowledge
\\
Multimodal Large Language Models (\S\ref{subsec:multimodal})  & Understanding content across multiple modalities using a single model
\\
Visual Generative Models (\S\ref{subsec:generative}) & Generate images with multiple styles or achieve image-style transformation
\\
\bottomrule
\end{tabular}
}
\label{tab:application_summary_lm}
\vspace{-10pt}
\end{table*}

\subsection{Model Merging in Large Language Models (LLMs)}
\label{subsec:llms}

\begin{figure*}[t]
\centering  
\includegraphics[width=1.0\textwidth]{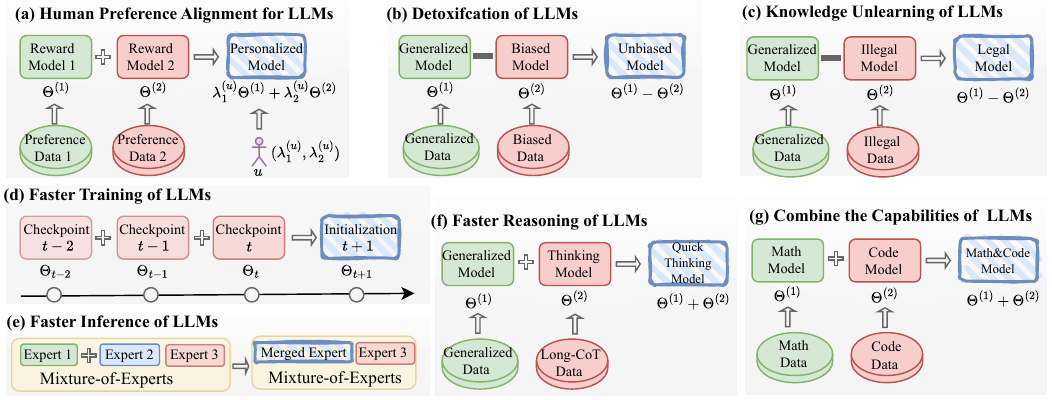}
\vspace{-20pt}
\caption{
\revised{Example application scenarios of model merging in large language models.}
}
\vspace{-15pt}
\label{fig:LLM_application}
\end{figure*}

In recent years, LLMs (such as GPT~\cite{gpt4}, Gemini~\cite{gemini}, PaLM~\cite{palm} and LLaMA~\cite{llama}) have made significant advancements and have been widely applied across various tasks. Despite their superhuman performance on most basic tasks, LLMs still face numerous challenges, including producing toxic content that violates laws or ethics, using unauthorized data during training, high training/inference/reasoning costs, and insufficient performance in specific domains. As shown in Figure \ref{fig:LLM_application}, model merging technology presents a promising opportunity to address these challenges.

\subsubsection{Human Preference Alignment for LLMs}
\label{subsubsec:llm_Alignment}

Humans often hold diverse opinions about aesthetics, politics, or fairness. When LLMs serve humans, different people have different expectations of the model, e.g., some expect LLMs to generate harmless responses, while others seek engaging and enjoyable interactions~\cite{rame2023rewarded}. Consequently, the development of practical LLMs is generally divided into three stages, to generate responses that are more helpful, accurate, and safer~\cite{lu2024online}: Pre-training on a large amount of unsupervised data, supervised fine-tuning (SFT) on a small dataset with high-quality annotation, and interaction with humans to further optimize LLM alignment (e.g., direct preference optimization (DPO)~\cite{rafailov2023DPO} or reinforcement learning from human feedback (RLHF)~\cite{RLHF2019}) with human preferences, rewards or values.

Some works propose to achieve better, safer, or faster alignment of human preferences by model merging~\cite{zheng2024weak,wang2024conditioned,bhardwaj2024language,fu2024disperse,pentyala2024paft}. For example, ExPO~\cite{zheng2024weak} adds a task vector, constructed by a moderate model aligned using DPO or RLHF on a small amount of human preference data, to an unaligned SFT model. A more powerful aligned model can be directly obtained by setting a suitable merging coefficient. On the AlpacaEval 2.0 benchmark~\cite{li2023alpacaeval}, fusing a model aligned on the 10\%/20\% preference data with an SFT model results in performance comparable to that of a model aligned on the full preference data. DogeRM~\cite{lin2024dogerm} proposed merging the reward model with LLMs fine-tuned on different downstream domains to create domain-private reward models directly. Additionally, \citet{lu2024online} propose an Online Merging Optimizer, that interpolates the gradient with the SFT model at each step of RLHF. This approach encourages RLHF to optimize toward reward maximization while preventing LLMs from forgetting general knowledge due to RLHF.
Beyond preference alignment, several studies have examined the impact of model merging for secure alignment of LLMs~\cite{bhardwaj2024language,yi2024safety,hammoud2024model,hazra2024safety}. For example, \citet{hammoud2024model} find that merging two security-aligned models could compromise security. Thus, they proposed explicitly including secure alignment as an optimization objective when constructing synthetic data for model merging.

In practice, users often have various combinations of preferences rather than a single preference. Training a model separately for each combination of preferences is unrealistic due to the infinite combinations and the high training costs. Therefore, some studies suggest combining models with different reward alignments to create a series of integrated aligned LLMs. 
For example, \citet{rame2023rewarded} and \citet{jang2023personalized} propose Reward Soups and Personalized Soups, respectively, as efficient and flexible solutions for diverse rewards. Specifically, Rewarded Soups first trains an expert model for each reward and then linearly interpolates the weights of the experts to approximate the set of Pareto optimal solutions for various reward combinations. This approach is cost-effective, as it only requires training separate models for each reward to combine any variety of rewards.

\subsubsection{Detoxification of LLMs}
\label{subsubsec:llm_Detoxifcation}

LLMs have been widely noted for issues related to untruthfulness and toxicity in various applications~\cite{hu2024separate}, such as insults, threats, and profanity in responses to certain questions. To address the potential security risks in the application of LLMs, flexible techniques are needed to reduce the generation of toxic text, essentially detoxifying LLMs. A straightforward solution is to collect additional non-toxic data to fine-tune LLMs~\cite{keskar2019ctrl}; however, this approach requires significant computing resources and may interfere with the general capabilities of LLMs. Alternatively, directly reducing the probability of potentially toxic words during the decoding stage requires additional guidance information~\cite{krause2021gedi}.
Recent studies have shown that reducing the toxic data generation of LLMs through model merging is a simple and effective scheme~\cite{TaskArithmetic_ICLR2023,pem_neurIPS2023,hu2024separate,dige2024mitigating}.

Task Arithmetic~\cite{TaskArithmetic_ICLR2023} negates the task vectors of GPT-2 model~\cite{gpt2_2019} fine-tuned on toxic data (Civil Comments~\cite{CivilComments_WWW2019}) and shows that this operation effectively reduces the proportion of data classified as "toxic", with little change in the fluency of the language on the control task (WikiText-103).
Additionally, some parameter-efficient models steer the toxic behavior of LLMs by manipulating a small number of parameters. PEM~\cite{pem_neurIPS2023} negates LoRA~\cite{lora_iclr2022} (and (IA)$^3$~\cite{IA3_NeurIPS2022}) modules trained on poisoning data to maintain language proficiency while reducing the toxicity of language model output. Ethos~\cite{gao2024ethos} and Ext-Sub~\cite{hu2024separate} point out that while the task vector on toxic data is factually wrong, it also contains correct information about language modeling and logical narrative skills. Therefore, Ext-Sub decomposes the toxic task vector into two orthogonal subspaces that represent general capability and destructive capability, respectively. Toxic knowledge is then eliminated by removing only the component representing the destructive ability from the LLM.

\subsubsection{Knowledge Unlearning of LLMs}
\label{subsubsec:llm_Unlearning}

LLMs may inadvertently learn copyrighted material, raising significant legal and ethical concerns~\cite{abad2024strongCopyright}, and broader questions about responsible AI use~\cite{dou2024avoiding}. In this context, the California Consumer Privacy Act~\cite{pardau2018california} and the General Data Protection Regulations of the European Union~\cite{hoofnagle2019european} stipulate the right to data forgetting. The foundational model's knowledge must be adapted to comply with these regulations. However, the cost of excluding copyrighted data for retraining from scratch is prohibitive. For instance, training a Llama-2-70B from scratch requires 1,720,320 GPU hours~\cite{touvron2023llama2}.
Traditional methods often use gradient ascent (GA) to achieve forgetting by fine-tuning the model using the GA algorithm on the specific data to be forgotten~\cite{thudi2022unrolling,yao2023large}. Unfortunately, this approach typically catastrophically destroys other parts of the model's knowledge. That is, forgetting specific knowledge also erases other knowledge that should be retained. Recently, many studies based on model merging techniques have demonstrated the potential to forget LLM-specific knowledge without harming other knowledge~\cite{TaskArithmetic_ICLR2023,dou2024avoiding, hu2024separate}. 

Unlike the GA-based approach, the model merging approach does not require additional data for other tasks to maintain old knowledge. To achieve forgetting, model merging typically incorporates a negatively fine-tuned model into the target model (i.e., the task-specific fine-tuned knowledge is subtracted from the target model). For example, 
Task Arithmetic~\cite{TaskArithmetic_ICLR2023} shows that negating task vectors degrades performance on specific tasks without substantial changes to the control tasks. Experiments demonstrate that model merging can forget the knowledge of the target task in a fine-tuned model without harming performance on control tasks. Similarly, Stable Sequential Unlearning~\cite{dou2024avoiding} extends this forgetting to the setting of sequential unlearning on LLMs, where different copyrighted content must be unlearned at different time steps.
Knowledge forgetting can also forget samples that represent bad behavior during pretraining. For instance, FuseToForget~\cite{zaman2023fuse} employs model merging as a debiasing tool to reduce privacy issues in language models. FLearning~\cite{ni2023forgetting} first subtracts the parameters related to the data to be forgotten and then fine-tunes the parameters with new data to achieve accurate knowledge updates. SKU~\cite{liu2024towards} explores the forgetting of harmful data in LLM, which is a two-stage scheme. Initially, harmful data (e.g., harmful question-answer pairs) is used to fine-tune the parameters corresponding to the location of harmful knowledge in the LLM (i.e., the task vector), and then the task vector is negated from the LLM to mitigate undesirable behavior in the LLM effectively. Generally, incorporating the opposite (anti-expert) task vectors into the pre-trained model can effectively accomplish the task of machine unlearning.

\subsubsection{Faster Training of LLMs}
\label{subsubsec:llm_Faster}

The training of LLMs requires numerous iterations on massive data, making the training process extremely expensive~\cite{kaddour2022stop,ram2024distribution}. For example, training LLAMA2-70B with 2T tokens required 1,720,320 GPU hours~\cite{liu2024checkpoint}. Methods to accelerate LLM training include mixed-precision training, continual retraining, and pipeline parallelism. An orthogonal approach is checkpoint merging in training trajectories, which offers a simple and effective means to either speed up LLM training or enhance training performance at the same cost.
Some works incorporate checkpoints in a single training trajectory during LLM training to accelerate model training. For instance, LAWA~\cite{kaddour2022stop} demonstrated that merging checkpoints during intermediate stages of model training speeds up the process. For example, training a RoBERTa-Base model on the WikiText-103 dataset saved 30 GPU hours. \citet{sanyal2023early} further showed that the combination of checkpoint averaging in pre-trained trajectories and a high learning rate contributes to faster convergence. Checkpoint Merging~\cite{liu2024checkpoint} comprehensively evaluates the effectiveness of model merging at different stages of the Baichuan2~\cite{yang2023baichuan} pre-training process. \revised{PMA~\cite{li2025model} and WSM~\cite{tian2025wsm} show that applying checkpoint averaging during LLM pre-training can serve as an effective alternative to learning rate decay schedules, while substantially reducing training costs.}

\subsubsection{\revised{Faster Inference of MoE-based LLMs}}
\label{subsubsec:llm_FasterInference}

\revised{
As the parameter scale of LLMs continues to grow, their performance improves significantly but often at the cost of substantially increased computation. In addition to dense LLMs, mixture-of-experts (MoE) architectures have recently attracted considerable attention for reducing inference costs (e.g., DeepSeek-MoE~\cite{dai2024deepseekmoe}, Mixtral 8$\times$7B~\cite{jiang2024mixtral}). MoE models lower computation by activating only a small subset of experts (e.g., two experts) for each input. However, using too few experts is efficient but degrades performance, whereas activating too many experts improves performance but increases cost.
Recently, model merging has been introduced to MoE architectures as a way to enjoy the benefits of using more experts without incurring a large computational overhead~\cite{he2023merging,li2025sub,miao2025mergemoe,nguyen2025expert}. For example, MEO~\cite{he2023merging} merges the parameters of multiple experts first and then computes the output based on the merged experts, instead of aggregating the outputs of all individual experts for each input. Sub-MoE~\cite{li2025sub} groups experts and performs SVD-based merging within each group. NAMEx~\cite{nguyen2025expert} further incorporates Nash bargaining into the merging process to better balance the contributions of different experts.
}

\subsubsection{\revised{Faster Reasoning of LLMs}}
\label{subsubsec:llm_FasterReasoning}

\revised{
Reasoning-oriented LLMs (e.g., DeepSeek-R1~\cite{guo2025deepseek}, QwQ-32B-Preview~\cite{team2025qwq}) have achieved remarkable improvements on complex tasks by engaging in iterative, chain-of-thought reasoning. However, this enhanced reasoning capability comes at the cost of significantly lower efficiency compared with conventional LLMs, and in some cases, the model may overthink even very simple problems, consuming a large number of tokens. Prior work~\cite{chen2024not} reports that QwQ-32B-Preview uses 901 tokens to answer a simple question such as “2+3=?”, whereas conventional models like Llama-3.3-70B and GPT-4o require only 7 tokens, and Gemini Pro only 5 tokens. Recently, a line of work has begun to reduce token consumption by merging “slow-thinking” reasoning models with “fast” conventional LLMs~\cite{wu2025revisiting,kimiteam2025kimik15scalingreinforcement,wu2025unlocking}. For example, the Kimi team~\cite{kimiteam2025kimik15scalingreinforcement} finds that merging long-context and short-context models is highly effective in improving token efficiency.
\citet{wu2025unlocking} conduct a detailed evaluation of different model merging methods and show that model merging provides an efficient mechanism for implementing long-to-short reasoning. Task-vector-based approaches, such as Task Arithmetic and TIES-Merging, can reduce inference length / token consumption by around 50\% on DeepSeek-R1-7B and Qwen2.5-Math-7B models while maintaining comparable accuracy.
}

\subsubsection{Combine the Capabilities of Expert LLMs}
\label{subsubsec:llm_CombineExpert}

LLMs exhibit strong generalizability in general tasks, but often lack knowledge in specific vertical domains. Pretrained LLMs typically require fine-tuning within different corporations to become expert LLMs in various fields. Integrating the expertise of multiple specialists is particularly critical for solving more complex tasks~\cite{tam2024llm}. Research on model merging techniques indicates that a composite LLM can be created by combining the parameters of different expert LLMs~\cite{DARE_Arxiv2023,akiba2024evolutionary,dekoninck2024controlled,zhou2024metagpt,fusionllm,wan2024fusechat,yu2024extend,sukhbaatar2024branch,yu2024extend,li2024s,ParamDelta}.
For example, \citet{dekoninck2024controlled} demonstrate the ability to flexibly control text generation by merging multiple LLMs with different styles and applying personalized weighting. Robust Weight Signatures~\cite{cai2023robust} proposes a robustness “patching” framework via model merging to enhance the overall robustness of the model against various naturally corrupted versions of clean data. 
\revised{Kwai Keye-VL 1.5~\cite{yang2025kwai} leverages model merging to combine domain-specific expert models with a LongCoT cold-start model into a unified general model, leading to improved performance. Souper-Model~\cite{maiti2025souper} demonstrates that model merging can improve both performance and stability across multiple capabilities, including multilingual understanding, tool calling, and mathematical reasoning, and achieves state-of-the-art results on the Berkeley Function Calling Leaderboard~\cite{patil2025the}.}
In summary, model merging offers a straightforward and effective strategy for enhancing LLM's capabilities.

\subsection{Model Merging in Multimodal Large Language Models (MLLMs)}
\label{subsec:multimodal}

Foundation models often involve processing and interacting with data from different modalities, such as video, images, speech, and text. In order to build a generally large model, a key obstacle is the diversity and heterogeneity of tasks and modalities. Traditionally, most existing approaches train a modality-specific model for each modality. However, these methods face limitations: on the one hand, they require separate models for each modality; on the other hand, jointly training a large multimodal model necessitates the expensive collection of paired training data (image, text, video, speech) and the retraining of the entire model when a new modality is added.

\begin{figure*}[t]
\centering  
\includegraphics[width=1.0\textwidth]{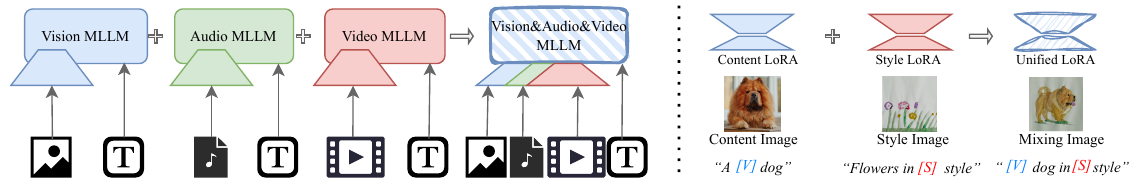}
\vspace{-15pt}
\caption{
Example application scenarios of model merging in multimodal large language models (left) and visual generative models (right).
}
\vspace{-10pt}
\label{fig:MLLM_application}
\end{figure*}

An interesting question is whether we can merge multiple modality-specific models to obtain a single, effective, and parameter-efficient modality-agnostic model. As shown in Figure \ref{fig:MLLM_application}(left), we aim for the merged unified model to encode inputs from different modalities, learn cross-modal interactions, and maintain performance comparable to that of well-trained independent modality-specific models. Compared to traditional multimodal learning, model merging techniques offer new opportunities. This model-merging approach offers several benefits: (1) it eliminates the costly and labor-intensive process of collecting labeled paired multimodal training examples, which is required for jointly training multimodal models; (2) it enhances the adaptability of multimodal models, allowing for the seamless integration of new modalities; and (3) it fully leverages knowledge collaboration across multiple modalities, thereby benefiting from cross-modal knowledge transfer.

\subsubsection{Model Merging for Multimodal Fusion}
\label{subsubsec:multimodalfusion}

Recently, many studies have focused on merging models from different modalities into a single model, thereby enhancing the diversity of knowledge across modalities. For instance, JAM~\cite{aiello2023jointly} proposes to merge two specialized (one for text-to-image and one text-only) autoregressive, decoder-only, large transformer models to seamlessly generate multimodal outputs. Similarly, DAMC~\cite{chen2024model} introduces a method for fusing multimodal LLMs across image, audio, video, and point cloud modalities, further reducing cross-modal interference through parameter decoupling and adjusting modality fusion coefficients.

To evaluate the impact of various factors on model merging, VL-Merging~\cite{MultimodalModelMerging_EMNLP2023} performs a comprehensive empirical analysis of multimodal model merging. The overall framework consists of three steps: independent modality fine-tuning, multimodal merging, and downstream task fine-tuning. Through experiments involving different initializations, merging methods, and architectures in multimodal model merging, the authors propose the following guidelines: (1) Models across multiple modalities should be based on the same pretraining starting point to ensure they are in the same basin~\cite{GitReBasin_ICLR2023} and share more information. (2) Simple model averaging achieves better performance, and if more computation and storage resources are available, more fine-grained merges can be conducted. (3) Merging the entire model rather than just a subset of layers generally yields more satisfactory results, as fine-tuning only a subset of layers may restrict the capabilities of single-modality models.

Unlike the above model merging approaches that are developed based on specific architectures, UnIVAL~\cite{shukor2023unival} is the first to design a unified architecture for four modalities: image, video, audio, and language. It transforms the tasks across all modalities into a "sequence-to-sequence" format, with the training objectives of all modalities converted into a "next token prediction" format. This allows for a uniform feature extraction and classifier to be applied across all modalities. Additionally, UnIVAL provides favorable architectural conditions for model merging and demonstrates that linear interpolation of models fine-tuned across multiple modalities in weight space results in a general single model that performs well on both seen and unseen tasks.

\subsubsection{Model Merging for Cross-Modal Knowledge Transfer}
\label{subsubsec:cross-modal}

Some works attempt to transfer knowledge from one modality to another through a model merging approach. For instance, 
MAM~\cite{sundar2023multimodal} investigates whether the attention layers of Transformers~\cite{transformer_2017} generalize across different modalities. Specifically, it examines if the knowledge acquired by Transformer models trained on high-resource modalities (e.g., data-rich images and text) can be transferred to Transformer models trained on low-resource modalities (e.g., data-sparse speech and audio). This paper demonstrates attention merging for models across various tasks, modalities, and initializations. The final results show that MAM achieves an 18.42\% reduction in classification error on the audio classification task (using the ESC-50 dataset~\cite{piczak2015esc}) compared to the standard fine-tuning paradigm.

\subsection{Model Merging in Visual Generative Models}
\label{subsec:generative}

The goal of visual generative models, such as generative adversarial networks (GANs) and denoising diffusion probabilistic models (Diffusions), is to approximate the underlying data distribution behind a given dataset so as to generate more new samples with the same distribution. However, image generative models still face the following challenges: the inability to flexibly generate samples with multiple style combinations and the high cost of generative model training.
This dilemma has led to an interest in expert models, which train a set of experts with specific abilities on different data shards or distributions, allowing for the flexible addition or removal of certain styles of experts at inference time. Considering the difficulty in deploying and the cost of resources for ensemble learning, model merging offers a new perspective on combining skill-specific experts of different styles without additional memory and inference costs.

\subsubsection{Style Mixing in Generative Models}
\label{subsubsec:generative_stylemix}
Existing generative models typically generate distributions based only on the training data. However, in real deployments, different users or artists often want to generate artwork with different combinations of styles. Collecting additional data for these mixed distributions is expensive, and fine-tuning the model can result in the forgetting of other capabilities. As shown in Figure \ref{fig:MLLM_application}(right),  model merging offers the potential to flexibly combine multiple styles.
Earl GAN Cocktail~\cite{avrahami2022ganCocktail} attempted to merge several pre-trained GAN models. Recently, diffusion-based image generative models~\cite{ho2020denoisingdiffusion,stablediffusion_2022} have gained more attention than GAN-based models due to their superior generative capabilities. Consequently, most research focuses on fusing different diffusion models.
Specifically, Diffusion Soup~\cite{DiffusionSoup2024} demonstrates the ability to linearly merge diffusion models fine-tuned on data shards of different styles (e.g., data provided by different domains/categories or different users), resulting in hybrid style zero-shot generation.  
In addition, Diffusion Soup empirically verifies that model merging has an anti-memorization effect, meaning the generated images are less likely to replicate training data, which is beneficial for generating diverse images. Unlike Diffusion Soup, which directly merges model parameters, MaxFusion~\cite{nair2024maxfusion} is inspired by ZipIt~\cite{ZipIt_Arxiv2023} and proposes merging intermediate features of multiple diffusion models based on the same input noise to generate images that satisfy multiple conditions.
However, merging multiple diffusion models based on full parameter fine-tuning can be costly when the number of tasks is large. To address this issue, ZipLoRA~\cite{shah2023ziplora} and MoLE~\cite{wu2024mole} aim to seamlessly merge parameter-efficient LoRA modules. For example, ZipLoRA proposes merging independently trained content/subject (e.g., a specific object or person) LoRAs with artistic style (e.g., drawing or painting, etc.) LoRAs, allowing the diffusion model to generate any user-provided combination of subjects and styles~\cite{Mergingloras}. This approach enables users and artists to easily combine publicly available subjects and styles LoRAs of their choice.
\revised{
Beyond image generation, \citet{tian2025extrapolating} apply model merging to image-to-video generation, enabling synthesized video clips that not only exhibit natural motion but also preserve the original appearance of the input image. 
}

\subsubsection{Reducing Training Cost of Generative Models}
\label{subsubsec:generative_cost}

In real-world scenarios, large-scale training data typically originates from different domains or is provided by various users. Given the need to add new data or remove outdated data, retraining a single model with updated data every time is often impractical~\cite{DiffusionSoup2024}. For instance, training a CM model~\cite{song2023consistency} using 8 A100 GPUs~\cite{lcsc_arxiv2024} takes about one week.
This is because existing methods only apply the final convergence weights during generative model training and ignore the intermediate training trajectories~\cite{thakur2024unified,lcsc_arxiv2024}. LCSC~\cite{lcsc_arxiv2024} demonstrates that a simple combination of training trajectories in the middle of the diffusion model by the evolutionary algorithm can significantly reduce the training cost. Specifically, only a few iterations or a small batch size is required to train the diffusion model to achieve image quality comparable to that of a fully trained diffusion model. For example, on the CIFAR-10 dataset, LCSC improves the training process for consistency distillation and consistency training~\cite{song2023consistency} by factors of $\small 23\times$ and $\small 7\times$, respectively. The underlying reason is that each local checkpoint of the optimized trajectory has many high-quality basins (i.e., areas of better generation quality) nearby that cannot be reached by stochastic gradient descent due to substantial variance in gradient estimations. However, checkpoint interpolation provides an opportunity to reach these basins.

\section{Application of Model Merging in Different Machine Learning Subfields}
\label{sec:application_ml}

Model merging is a simple and effective technique widely used in various subfields of machine learning, such as continual learning, multi-task learning, domain generalization, federated learning, few-shot learning, and adversarial defense, etc. In this section, we comprehensively discuss the application of model merging in the different machine learning subfields.
Table~\ref{tab:application_summary_ml} provides a brief summary, and in \S\ref{subsec:continual} to \S\ref{subsec:adversarial}, we introduce each application case in detail.

\begin{table*}[h]
\small
\caption{A summary of the application of model merging techniques in different machine learning subfields.}
\vspace{-10pt}
\resizebox{\linewidth}{!}{ 
\begin{tabular}{c|c}
\toprule  
\textbf{Scenarios} &  \textbf{The Main Purpose of Model Merging}    
\\ \midrule
Continual Learning (\S\ref{subsec:continual}) & Avoiding catastrophic forgetting with respect to old tasks
\\
Multi-Task / Multi-Domain / Multi-Objective / Auxiliary Learning  (\S\ref{subsec:multitask}) & Performing multiple tasks / domains / objectives via one model
\\
Domain / Out-of-Distribution Generalization (\S\ref{subsec:oodg_dg}) & Achieving generalization to unknown target domains or distributions
\\
Federated Learning (\S\ref{subsec:federated}) & Merging local models provided by different clients 
\\
Zero-shot / Few-shot Learning  (\S\ref{subsec:fewshot}) & Multiple related models are merged to improve the zero-shot / few-shot learning ability on new tasks
\\
Adversarial Learning (\S\ref{subsec:adversarial}) & Implementing model poisoning attack, defense, and copyright protection
\\
\bottomrule
\end{tabular}
}
\label{tab:application_summary_ml}
\vspace{-10pt}
\end{table*}

\subsection{Model Merging in Continual Learning}
\label{subsec:continual}

Continual Learning (CL) involves training a model using a streaming, non-stationary data stream. The primary challenge in CL is the `catastrophic forgetting'~\cite{catastrophicforgetting_1995,wang2023comprehensive} problem; that is, the CL model's prediction accuracy for old tasks drops dramatically after training on new tasks. The mainstream CL methods are mainly divided into memory replay-based methods, architecture expansion-based methods, regularization-based methods, and subspace projection-based methods~\cite{wang2023comprehensive}. Recently, there has been a growing interest in using model merging to address the catastrophic forgetting problem. This novel approach offers several benefits, such as avoiding additional parameters and inference costs associated with network expansion-based methods and eliminating the need to cache old data as required by memory-based methods.

\label{subsubsec:forgetting}

To overcome catastrophic forgetting and maintain new learning ability, many works have proposed to merge models for old tasks and update models for new tasks~\cite{wortsman2022robust,schumann2024backward,panda2024lottery}.
\revised{
For example, WiSE-FT~\cite{wortsman2022robust} shows that merging the fine-tuned model with the pretrained model can effectively mitigate the forgetting of general knowledge.
}
Tangent Model Composition~\cite{liu2023tangent_tmciccv2023} proposes fine-tuning each task independently in the tangent space of the pre-trained model and then merging the resulting models to perform CL.
In addition, ITA~\cite{porrello2024second} emphasizes the necessity for the fine-tuned model to be in the same basin as the pre-trained model to ensure the composability of nonlinear models. It introduces a regularization term similar to EWC~\cite{kirkpatrick2017overcoming} in traditional CL to constrain the distance between the fine-tuned weights and the pre-trained weights when training the independent model. WARP~\cite{rame2024warp} suggests linearly interpolating the pre-trained LLM's weights with its aligned weights via RLHF on a preference dataset, thus mitigating the forgetting of knowledge from the pre-trained LLM. BAM~\cite{alexandrov2024cfllm} continuously adapts LLMs to new languages by merging models while preserving general capabilities. Model Tailor~\cite{zhu2024model} explores the problem of catastrophic forgetting during fine-tuning of MLLMs, and proposes to merge only the most important subset of parameters in the fine-tuned MLLM model into the pre-trained MLLM model, so as to retain the generalization ability of the pre-trained model as much as possible, while compensating the selected weights to reduce the performance of the fine-tuning task. MagMax~\cite{marczak2024magmax} merges pruned task vectors to further alleviate parameter sign conflicts and old knowledge forgetting.
Equifinality, PAINT~\cite{ilharco2022patching} and LM-Cocktail~\cite{Lmcocktail} interpolate the weights of the fine-tuned model and the zero-shot model to improve accuracy on downstream tasks without degrading accuracy on supported/general tasks.

In contrast to merging full models, some research focuses on merging parameter-efficient modules. 
\citet{chitale2023task} propose a CL method based on task arithmetic~\cite{TaskArithmetic_ICLR2023}.  This method first fine-tunes a task-specific LoRA for each task, then constructs a task vector based on the difference between fine-tuned and pre-trained models. Multiple task vectors are then merged, and a small amount of data (10 samples per class) is used to fine-tune the merged model. Compared to traditional CL methods, particularly those based on replay, this approach eliminates the need to replay data from old tasks at each iteration, thereby accelerating model training. Additionally, fine-tuning the merged model with a class-balanced subset helps mitigate CL model bias. Similarly, DynaMMo~\cite{qazi2024DynaMMo} applies lightweight model merging (i.e., Adapter) in a CL setting for medical images. In contrast to architecture expansion-based CL methods, this approach does not result in a linear increase in the number of parameters with the number of tasks. Unlike the static aggregated parameter-efficient fine-tuning (PEFT) modules of DynaMMo, DAM~\cite{cheng2024dam} introduces dynamic aggregated PEFT modules during inference to perform CL.
AMM~\cite{chen2024adaptive} proposes merging convolutional layers to facilitate incremental new class discovery and prevent forgetting fundamental knowledge.  Disperse-Then-Merge~\cite{fu2024disperse} suggests merging submodels trained on different data partitions during the supervised fine-tuning of LLMs to reduce data bias and mitigate the forgetting of generic pre-trained knowledge.

\subsection{Model Merging in Multi-Task/Multi-Objective/Multi-Domain/Auxiliary Learning}
\label{subsec:multitask}

In machine learning, to optimize resource efficiency, we typically use a single model to handle multiple tasks, objectives, or domains with varying distributions. The traditional multi-task learning (MTL), multi-objective learning (MOO), or multi-domain learning (MDL) paradigm requires gathering data from all tasks, objectives, or domains to collaboratively train a model, leading to high data management and model training costs. This approach becomes particularly costly when new tasks, goals, or domains are introduced, as retraining a comprehensive model from scratch using all available data is resource-intensive. Numerous recent studies have proposed efficient methods for integrating knowledge across tasks, goals, or domains by merging models directly.

\subsubsection{Knowledge Transfer in Multi-Task Learning}
\label{subsubsec:kt_mtl}

The goal of MTL is to enable a single model to perform multiple tasks simultaneously, thereby facilitating knowledge transfer between these tasks~\cite{mmoe_kdd2018,gradnorm_icml2018,mtlasmooSenerK18_neurips2018,Adashare_NeurIPS2020,PCGrad_NeurIPS2020,adatask_AAAI2023}. As shown in Figure~\ref{fig:same_architectures}(b), to avoid the high cost of joint training, a straightforward approach is to merge multiple independently trained models on different tasks to accomplish MTL. Almost all of the model merging methods discussed in \S\ref{subsec:duringmerging} can be used to merge multiple models trained on different tasks to perform MTL. In this section, we take some representative tasks as examples. For MTL tasks in computer vision, Task Arithmetic~\cite{TaskArithmetic_ICLR2023}, Ties-Merging~\cite{TiesMerging_NeurIPS2023}, AdaMerging~\cite{AdaMerging_ICLR2024} and other studies~\cite{ye2023merging,tang2023concrete,surgery_icml2024} proposed to combine ViT models trained on different visual classification tasks, and the obtained model can complete the object classification of multiple tasks. The results of Task Arithmetic~\cite{TaskArithmetic_ICLR2023} demonstrate that merging independently trained models from any two datasets yields a merged model whose performance is comparable to that of a single-task model. Similarly, ZipIt~\cite{ZipIt_Arxiv2023}, which merges ResNet architectures trained on different tasks, achieves comparable results.
For MTL tasks in natural language processing, DARE~\cite{DARE_Arxiv2023} introduces a method to assimilate homologous models, augmenting LLMs as a "free lunch". For instance, merging WizardLM with WizardMath significantly boosts WizardLM's performance on GSM8K (a benchmark for evaluating the mathematical reasoning ability of LLMs) from 2.2 to 66.3.  \citet{akiba2024evolutionary} suggest that directly merging an LLM with mathematical capabilities and an LLM with Japanese language proficiency results in a model capable of solving Japanese mathematical problems.
Furthermore, numerous studies have demonstrated that combining PEFT modules trained on different tasks can also achieve MTL~\cite{LinearizationLoRA_ICLR2024,pem_neurIPS2023,belanec2024task}.

\subsubsection{Knowledge Transfer in Multi-Objective Optimization}
\label{subsubsec:kt_moo}

MOO aims to optimize multiple objective functions simultaneously. These objective functions may conflict with one another, so the MOO problem typically does not have a single optimal solution. Instead, it involves finding a trade-off among the multiple objectives, which corresponds to identifying a set of Pareto optimal solutions.
\citet{tang2024towards} propose approximating the entire Pareto set using a mixture of experts based model merging approach. Specifically, their method trains an independent model for each objective and learns a routing network to balance the trade-offs between the multiple objectives. The input of the routing network is the task preference vector, and its output consists of the merging coefficients for the independent models. \revised{Pareto Merging~\cite{chenpareto} also formulates different preferences as an MOO problem, and generates a Pareto set of merged models.}
Considering that directly evaluating Pareto solutions based on the original evaluation metric is time-consuming, MAP~\cite{li2024map} proposes a second-order Taylor expansion model as a surrogate model for the true evaluation metric, and further uses an evolutionary algorithm to calculate the Pareto front based on the surrogate model.

\subsubsection{Knowledge Transfer in Multi-Domain Learning}
\label{subsubsec:kt_mdl}

Unlike existing model-merging-based MTL approaches that focus on datasets with different object categories, \citet{ye2023merging} explore model merging across multiple domains, where datasets share the same categories but differ in environmental contexts. To mitigate conflicts between multi-domain models, this paper introduces a weight similarity criterion to assess the correlation between different model layers. For layers with high correlation, a simple weight averaging or RegMean~\cite{RegMean_ICLR2023} strategy is employed to merge models that have been fine-tuned in different domains of the same task. For layers with low correlation, the weights are flexibly combined using a gating mechanism during the inference phase. Branch-Train-Merge~\cite{li2022branch} demonstrates the effectiveness of training expert language models on 64 different domains and subsequently merging them.

\subsubsection{Knowledge Transfer in Auxiliary Task Learning}
\label{subsubsec:kt_al}

The goal of ATL is to enhance the performance of the target task by leveraging knowledge obtained from related auxiliary tasks. Unlike MTL, which aims to optimize the average performance across all tasks, ATL focuses solely on improving the performance of the main task. However, ATL often encounters the issue of gradient conflict, leading to negative transfer, where the inclusion of auxiliary tasks interferes with the main task's performance. To mitigate negative transfer, \citet{jiang2024forkmerge} propose ForkMerge, which periodically performs `fork' and `merge' operations. The model is first periodically duplicated into multiple branches: the first branch is trained exclusively on the main task, while the remaining branches are trained jointly on both the main and auxiliary tasks. An optimal merging coefficient is then determined using the validation set to merge the models updated by the various branches. Empirical results show that ForkMerge achieves positive transfer gains across several ATL benchmarks.

\subsection{Model Merging in Out-of-Distribution/Domain Generalization}
\label{subsec:oodg_dg}

The common goal of out-of-distribution generalization (OODG) and domain generalization (DG) is to improve a model's performance on unseen data. The key difference between them is that OODG focuses on enhancing a model's generalization ability on unknown data with significantly \textit{different distributions} from the training data. In contrast, DG emphasizes improving a model's generalization ability on \textit{unseen domains}. Numerous recent studies have demonstrated that model merging contributes to enhanced training stability and overall performance in both OODG and DG.

\subsubsection{Model Merging for Better Out-of-Distribution Generalization}
\label{subsubsec:oodg}

In real-world scenarios, a trained model may be deployed in environments with changing distributions. For example, autonomous driving models are trained on a clean dataset, but in practice, they are vulnerable to unforeseen distributions such as natural corruptions (e.g., camera noise, motion blur) and more significant distribution shifts (e.g., summer to winter)~\cite{hendrycks2019robustness,cai2023robust}. The goal of OODG is to enhance the model's ability to generalize to unknown data that significantly differs from the training distribution.

Stochastic weight averaging (SWA)~\cite{izmailov2018averaging} is a straightforward and widely used technique to improve machine learning models' training stability and OOD performance. From a statistical perspective, weight averaging helps reduce variance during model training. Many works merge intermediate weight states (i.e., checkpoints) from training trajectories while training models~\cite{tarvainen2017mean,izmailov2018averaging,yang2019swalp,SWAP_ICLR2020,zhang2023lookaround,von2020neural}. For example, WiSE fine-tuning~\cite{wortsman2022robust} demonstrates that linearly combining the weights of a pre-trained model and a fine-tuned model can significantly improve accuracy in the case of distribution shifts, while maintaining high accuracy on the original distribution. SWA~\cite{izmailov2018averaging,SWAP_ICLR2020} simply averages all checkpoints from the beginning of a particular epoch to the end of training. This approach is explained to help the model converge to flat rather than sharp local optima, thereby improving generalization~\cite{izmailov2018averaging, kaddour2022flat}. Adaptive SWA~\cite{ASWA_JMLR2024} highlights that executing SWA too early may lead to underfitting, while executing it too late may result in overfitting. It proposes averaging only when generalization on the validation set improves, effectively combining SWA with an early stopping mechanism. TWA~\cite{TWA_ICLR2023} addresses this by showing that the averaging coefficients of the weights can be determined in a training manner. Consequently, TWA, unlike simple SWA, can perform averaging from the initial epoch of training, eliminating the need to define an additional hyperparameter for the epoch at which weight averaging should start.

In contrast to previous works that average weights obtained along one training trajectory, methods such as Model Soups~\cite{Modelsoups_ICML2022,SparseModelSoups}, AdapterSoup~\cite{adaptersoup_EACL2023}, Model-Ratatouille~\cite{rame2023modelratatouille}, WARM~\cite{rame2024warm}, WARP~\cite{rame2024warp}, PAPA~\cite{papa2024}, WASH~\cite{fournier2024wash}, DART~\cite{jain2023dart}, and DiWA~\cite{rame2022diverse} propose merging multiple independently fine-tuned or trained models. These models are usually more diverse, which improves OOD performance. Independently trained models differ in hyperparameters (e.g., learning rate, weight decay, dropout), batch order, data augmentation techniques (e.g., random crops, horizontal flips), and the number of training steps, among other factors. Specifically, Model-Ratatouille~\cite{rame2023modelratatouille}, starts from the same initial model, fine-tunes multiple models on an auxiliary task, then continues to fine-tune these models on the target task, and finally merges the diverse models to improve OOD performance. WARM~\cite{rame2024warm} further increases the diversity of fine-tuned models by sampling different checkpoints from the trajectories of the pre-trained model as the initial weights for the downstream preference fine-tuning task.  To reduce the additional cost of training multiple models, Model Stock~\cite{jang2024model} proposes that we can exploit the geometric properties of the weight space and the anchoring effect of pretrained models to approximate the merged weights using only a few fine-tuned models. MEHL-Soup~\cite{li2024scalable} develops a scalable and efficient method to learn merging coefficients for model soup. It only loads a subset of models for each iteration, significantly reducing the computation and memory requirements of naive model soup for learning merging coefficients.

The above analysis reveals that the SWA lacks diversity due to its reliance on a single trajectory. In contrast, Model Soups and DiWA train independently, which can lead to multiple models with significant differences, resulting in weight averaging failure. To balance these two approaches, Lookaround~\cite{zhang2023lookaround} introduces a gradient descent optimizer based on the concept of weight averaging. This optimizer iteratively performs `around' and `average' steps throughout the optimization process. In the `around' step, multiple independent models are trained from the same starting point, each using different data augmentations. In the `average' step, the diverse models are averaged, and the result is used as the starting point for the next iteration.

\subsubsection{Model Merging for Better Domain Generalization}
\label{subsubsec:dg}

Domain generalization methods aim to generalize to an \textit{unknown target domain} using only training data from source domains.  For instance, in the context of traffic sign recognition, the training data for a machine learning model tasked with identifying traffic signs in various urban environments comes from multiple cities (i.e., source domains). However, when deployed, the model must recognize traffic signs in new urban environments (i.e., target domains) that it has never encountered before. Existing DG methods can be classified into domain alignment, data augmentation, regularization, and meta-learning frameworks~\cite{arpit2022ensemble}. Complementary to these approaches, model merging techniques can be seamlessly integrated to further improve out-of-domain performance without modification~\cite{SWAD_NeurIPS2021,arpit2022ensemble,li2024training}. 
SWAD~\cite{SWAD_NeurIPS2021} demonstrates that flatter minima generalize better to unseen domains. Inspired by SWA~\cite{izmailov2018averaging}, SWAD proposes a dense and overfit-aware stochastic weight sampling strategy to identify these flatter minima. More specifically, unlike SWA, it starts from a predefined epoch until the final epoch, and collects a random weight every $K$ epochs for averaging. SWAD collects weights densely, that is, one is collected every step, and the start and end of random weight collection are determined by the performance changes on the validation set. EoA~\cite{arpit2022ensemble} also shows that model averaging can improve out-of-domain stability, and that ensembling multiple moving average models can further enhance performance compared to ensembling models without weight averaging.

\subsection{Model Merging in Federated Learning}
\label{subsec:federated}

Federated Learning (FL) is a distributed learning approach that allows multiple clients to collaboratively train a model without sharing data. FL primarily includes two settings: centralized (with a central server) and decentralized (without a central server). Each client updates the model or calculates the gradient based on local data and sends the updated information to the central server (in centralized FL) or other clients (in decentralized FL) for aggregation to update the global model, thus ensuring data privacy protection.
In FL, model merging refers to summarizing parameters from various clients during each communication round, thereby forming an updated global model.

\label{subsubsec:fl_local}
Most FL methods adopt a simple coordinate-wise average to aggregate the local models. For example, they calculate local model merging coefficients according to some heuristic rules. FedAvg~\cite{FedAvg_AISTAT2017}, the most classic FL method, proposes to merge local models on the server weighted by the amount of training data from each client. FedNova~\cite{Fednova_2020} normalizes and scales model updates on the client side based on the number of update steps, efficiently aggregating local models to obtain a high-performance global model. FedAtt~\cite{ji2019learning} calculates layer-wise attention coefficients based on the similarity of client and server parameters, fusing local models based on these coefficients. 
FedFisher~\cite{jhunjhunwala2024fedfisher} computes the Fisher information matrix of the parameters in each client to merge the local models.
In more challenging FL tasks, the above direct coordinate-wise merging methods may result in suboptimal global model performance. Inspired by the property of permutation invariance of neural networks, PFNM~\cite{yurochkin2019PFNM}, OTFusion~\cite{otfusion_neurips2020} and FedMA~\cite{FederatedMA_ICLR2020} propose to permute neurons of local models before merging them.
Similarly, GAMF~\cite{gamf_icml2022} transforms the merging problem into a multi-graph matching problem based on graph matching and then merges the aligned local models.

\subsection{Model Merging in Zero-shot/Few-shot Learning}
\label{subsec:fewshot}

In practical applications of machine learning models, collecting a large amount of labeled data can be expensive or infeasible in specific scenarios (e.g., medical diagnosis, real-time monitoring). Users often want models to effectively perform new tasks that have not been encountered before, that is, an ability commonly referred to as \textit{cross-task generalization}~\cite{LoraHub_Arxiv2023}. Zero-shot~\cite{mishra2022cross} and few-shot learning~\cite{ye2021crossfit} can reduce the dependence on large amounts of data and allow the model to better deal with unseen categories or small numbers of samples, improving the cross-task generalization ability of the model. In few-shot learning, a common approach is to fine-tune the model using the limited examples available. However, because of the minimal data, this fine-tuning process is often unstable and yields only modest performance improvements. Recently, some studies have explored merging pre-trained models (from some publicly accessible resources) to enhance cross-task generalization under zero-shot and few-shot conditions~\cite{wu2023pi,LoraHub_Arxiv2023,jang2023exploring,ostapenko2024towards,muqeeth2024learning,wang2024lora,tao2024unlocking}.

\subsubsection{Model Merging for Cross-task Generalization in Zero-shot Learning}
\label{subsubsec:kt_zeroshow}

Model merging has demonstrated the effectiveness of zero-shot learning across several applications. Some examples of practical applications include cross-lingual transfer~\cite{huang2024chat, chronopoulou2023language, zhao2024adamergex,klimaszewski2024no}, hybrid style image generation~\cite{DiffusionSoup2024, nair2024maxfusion}, and multi-modal processing~\cite{chen2024model}.
Some works achieve \textit{cross-lingual transfer} through model merging, such as chat~\cite{huang2024chat}, text summarization~\cite{chronopoulou2023language}, or reasoning~\cite{zhao2024adamergex}. A well-performing language-specific LLM needs to be fully trained, and with 7,000 languages in the world, not all of them have enough labeled data to support model fine-tuning. Therefore, cross-lingual knowledge transfer is particularly important.
For example, \citet{huang2024chat} build a Chat vector based on fine-tuned LLAMA2-chat and pre-trained LLAMA2 on chat data in the English language, and assembles it with the continuously pre-trained LLAMA2 model on other non-English languages. This allows the new model to chat in non-English languages.
\citet{chronopoulou2023language} develop a zero-shot multilingual summarization framework. It uses a merged model (a supervised summarization model and an unsupervised pre-trained model for a high-resource language, along with an unsupervised pre-trained model for a low-resource language) to perform text summarization tasks in low-resource languages. Similarly, AdaMergeX~\cite{zhao2024adamergex} demonstrates the effectiveness of model merging for cross-language transfer across three tasks: reasoning, natural language understanding, and natural language generation.
In the \textit{hybrid-style image generation task}, Diffusion Soup~\cite{DiffusionSoup2024} and MaxFusion~\cite{nair2024maxfusion} show that the zero-shot generation ability can be enhanced by merging multiple diffusion models.
In the \textit{multi-modality task}, DAMC~\cite{chen2024model} experiments prove that zero-shot multi-modal extension can be achieved by merging multi-modal models, provided they are initialized from the same LLM. For example, by merging a visual LLM and an audio LLM, the combined model can not only perform image or audio tasks independently but also acquire the zero-shot ability to process inputs containing both visual and auditory information simultaneously.

\subsubsection{Model Merging for Cross-task Generalization in Few-shot Learning}
\label{subsubsec:kt_fewshow}

Parameter-efficient fine-tuning (PEFT), such as LoRA or Adapter, facilitates the creation and sharing of thousands of custom PEFT modules, each trained on different data for various downstream tasks. A natural question is whether combining PEFT modules pre-trained on different upstream tasks can improve the transfer accuracy for unseen downstream tasks with limited samples. 
Recent work on model merging suggests a positive answer, showing that merged models can enhance generalization in few-shot settings~\cite{he2023mera,asadi2024does,LoraHub_Arxiv2023,wang2024lora}. For example, LoraHub~\cite{LoraHub_Arxiv2023} proposes to merge LoRA modules available on HuggingFace to achieve adaptive performance for unseen tasks, where the merging coefficients of different LoRA are searched in a black-box gradient-free manner with few-shot samples. As expected, few-shot LoraHub performs better than few-shot in-context learning and reduces inference costs by eliminating the need for examples as input to LLMs.  LoraRetriever~\cite{zhao2024loraretriever} further proposes dynamically retrieving the most relevant LoRAs based on the input and merging them. Similarly, MerA~\cite{he2023mera} proposes merging pretrained adapters into a single adapter for few-shot NLP scenarios. 
In general, well-trained LoRAs or adapters can serve as valuable resources that users can easily share, access, and apply to a variety of downstream tasks. In the real world, upstream and downstream tasks can be entirely disparate, originating from different datasets, domains, or even different parts of the same dataset. \citet{asadi2024does} comprehensively evaluates model merging in the few-shot learning setting. Specifically, this study examines three cases of label, domain, and task drift between upstream and downstream tasks. The results demonstrate that model merging enhances the model's generalization ability in few-shot learning scenarios across different contexts.

\subsection{Model Merging in Adversarial Learning}
\label{subsec:adversarial}

In the machine learning community, the open-source availability of pre-trained models (e.g., LLaMA~\cite{llama}) has accelerated technological advancements. In this context, developers often download unvalidated checkpoints to fine-tune their models or even outsource the training process to third-party platforms~\cite{backdoorbench2022}. Consequently, open-source models are also vulnerable to malicious attacks, such as poisoning attacks, where hidden malicious behaviors can be triggered by specific inputs. This raises several intriguing questions: Can model merging lead to attacks, and can it be used to develop defense mechanisms? Additionally, how can intellectual property protection be enhanced in the context of model merging?

\subsubsection{Model Merging as an Attack}
\label{sssec:attack}

Parameter-efficient fine-tuning  methods~\cite{ding2023parameter_nmi}, such as LoRA~\cite{lora_iclr2022}, exhibit functional transferability. This means that a LoRA model fine-tuned for a specific task based on a pretrained model can be successfully transferred to another pretrained model~\cite{liu2024lora}. In practice, developers often download LoRA models from open-source platforms to address their specific downstream tasks~\cite{LoraHub_Arxiv2023}. If a poisoned LoRA, which could be seen as a Trojan horse, is inadvertently downloaded and integrated into a model, it may introduce security vulnerabilities. Research by LoRA-as-an-Attack~\cite{liu2024lora} demonstrates that merging a poisoned LoRA—trained on compromised data—with a benign LoRA, trained on clean data, can result in a backdoor injection. This finding also holds when multiple LoRAs are merged. In addition, BadMerging~\cite{zhang2024badmerging} has developed a two-stage backdoor attack framework specifically for model merging, and through a large number of experiments, it has shown that the success rate of on-task and off-task attacks on merged models exceeds 90\%, and existing defense measures cannot defend against BadMerging.

\subsubsection{Model Merging as a Defense or Intellectual Property Protection}
\label{sssec:defense}

Contrary to the attacks described in \S\ref{sssec:attack}, the transferability of LoRA also offers an opportunity for model merging as a defense strategy. Specifically, if we know that a model may be susceptible to certain attacks, can we train some LoRAs to enhance the model's defense (i.e., reduce the attacker's success rate)? For example, \citet{liu2024lora} demonstrate that GPT-3.5 was used to generate a benign dataset containing backdoor triggers. A dedicated defense LoRA was then trained on this benign data and merged into the poisoned pre-trained model. This defensive model merging ultimately led to a reduction in the backdoor effect. Furthermore, research has shown that in the context of full parameter fine-tuning, model merging can serve as a "free lunch" for model defense. Experiments involving four model architectures and four datasets revealed that merging multiple poisoned models without additional effort mitigated these poisoning attacks, with the accuracy on the benign dataset remaining nearly unaffected. \citet{rebuffirevisiting} and \citet{croce2023seasoning} merge a set of $l_p$ (for various $p$) robust fine-tuned models to easily control the robustness level of each threat model against $l_p$ boundary adversarial attacks.
Similarly, the experimental analysis by \cite{jailbreaks2024} indicates that model merging offers an effective defense mechanism against jailbreak attacks.

In another practical scenario, merging unauthorized models may infringe on the intellectual property rights of the model owner. Malicious users might merge several high-quality open-source models (e.g., those authorized for research use only) to create a new model, then claim that this new model was entirely developed and trained from scratch by themselves, subsequently offering model services for commercial gain. In such cases, it becomes particularly crucial for model owners to detect whether others have merged their models. MergeGuard~\cite{cong2024have} performs a preliminary analysis of the effectiveness of two existing defense methods—Quantization Watermarking~\cite{li2023watermarking} and Instructional Fingerprint~\cite{xu2024instructional}—in the context of model merging. The study observed that while the watermarking method cannot be detected in the merged model, the fingerprint method remains detectable.

\section{Remaining Challenges and Future Directions}
\label{sec:future_directions}

Although \S\ref{sec:methods}--\S\ref{sec:application_lm} and \S\ref{sec:application_ml} present various advanced model merging methods and applications, challenges remain in the technology and application of existing model merging approaches. 
Additionally, there are numerous areas that warrant further research in the future.

\textbf{(1) Closing the Performance Gap Between the Merged and Independent Models}.
The effectiveness of current model merging techniques heavily relies on the "pretraining-finetuning" paradigm. Specifically, successful model merging requires that multiple models be fine-tuned based on the same pre-trained model, with careful control over the number of epochs and learning rate during fine-tuning. If these hyperparameters are not set properly, the models may not converge in the same or close basin. Even based on the pre-trained fine-tuning paradigm, there is still a significant gap between the merged and independent models, especially when the number of models/tasks is large. \revised{As shown in Figure \ref{fig:acc_performance}, when merging eight tasks, the independently fine-tuned ViT-B/32 model achieves an accuracy of 90.3, whereas the simplest weight averaging baseline reaches only 66.5. Meanwhile, for Llama-3.1, the best model merging method attains a score of 84.8, which still lags far behind the expert model’s score of 100.}
Therefore, a promising direction for future research is to explore how to ensure the effectiveness of model merging under more relaxed conditions. For example, investigating how to merge multiple models that are trained independently from scratch for different tasks without compromising performance could be valuable.

\textbf{(2) In-depth Theoretical Analysis for Model Merging}.
The validity and explanation of existing model merging techniques are largely empirical and lack sufficient theoretical guarantees. As discussed in \S\ref{subsec:theory}, there is currently a limited amount of work on the theoretical aspects or explanations of model merging. The few existing studies mainly focus on merging multiple models trained on the same trajectory or on the same dataset with different fine-tuning settings. There is almost no theoretical research or explanation concerning merging multiple models fine-tuned on different datasets or merging multiple models trained from scratch on different datasets. Therefore, future research should aim for a more comprehensive and in-depth theoretical analysis to enhance the success and reliability of model merging.

\textbf{(3) Trustworthy Model Merging}.
Model merging is prone to intellectual property disputes and poisoning attacks, making the development of a reliable and trustworthy merging scheme an urgent research priority. Research on the reliability of model merging can be categorized based on two key roles: the model owner and the model combiner.  
On one hand, for model owners, protecting the intellectual property of their models is a primary concern. This protection involves both active and passive defense strategies: (1) \textit{Active Defense}: Model owners may want to ensure that their published models are used independently and not merged by other users. The ideal outcome of an active defense strategy is that the model performs stably when used as intended, but breaks down completely if merged with other models. (2) \textit{Passive Defense}: When model owners suspect that their models have been merged, there needs to be a robust method to verify whether the merged models contain their original models.
On the other hand, for the model combiner, a key research direction is how to effectively prevent the inclusion of malicious injections, such as backdoors or poisoning attacks, when merging a set of authorized models.

\textbf{(4) Effective and Efficient Model Merging}.
Existing high-performance model merging methods often come with high costs in terms of efficiency and memory. First, most of these methods require all models to be loaded into memory during execution. For instance, merging 72 fine-tuned ViT-B/32 models necessitates more than 200GB of memory~\cite{li2024scalable, Modelsoups_ICML2022}. Additionally, heuristics for determining model merging coefficients involve repeated evaluations of the combined model, while learnable methods depend on additional data and training. In the future, it would be beneficial to develop more efficient model merging methods that do not require training, additional data, GPUs, or large amounts of memory.

\textbf{(5) Merge Heterogeneous Models}.
Existing methods primarily focus on merging homogeneous models. However, in practice, numerous heterogeneous models excel in various tasks. A limited number of existing methods for merging heterogeneous models involve transforming multiple heterogeneous models into homogeneous ones using the knowledge distillation technique~\cite{avrahami2022ganCocktail,wan2024fusechat}. The distillation process relies on the data from the original tasks and involves costly training. Therefore, it is also worth exploring approaches to merge these heterogeneous models without incurring the high costs associated with architectural transformations.

\textbf{\revised{(6) Develop Merger-Friendly Fine-tuning Strategies}}.
\revised{
Most existing model merging methods primarily focus on mitigating parameter conflicts and interference at the merging stage. Only a few works consider improving the fine-tuning procedure itself so that the resulting models are more amenable to merging, i.e., they incur smaller performance degradation compared with standard fine-tuning. Designing better fine-tuning strategies that facilitate subsequent merging, without significantly increasing computational cost or imposing strict constraints on developers, is therefore a highly meaningful direction for future work.
}

\textbf{(7) Interdisciplinary Application of Model Merging}.
As discussed in \S\ref{sec:application_lm} and \S\ref{sec:application_ml}, model merging has been adeptly applied across various foundation models and machine learning subfields to address different challenges and achieve interesting tasks. The question of how to adapt model merging strategies from one subfield to another presents an exciting avenue for exploration. In addition, it is important to further explore additional application scenarios for model merging.

\section{Conclusions}
\label{sec:conclution}

Model merging is a straightforward and effective technique for model enhancement that combines multiple models to achieve diverse capabilities. In this survey, we first provide a comprehensive overview of the advanced methods and theories currently available in the field of model merging. Next, we discuss the application of model merging techniques across various foundation models (i.e., LLMs, MLLMs) and more than ten subfields of machine learning, highlighting their use in addressing various challenges and difficulties. Finally, we identify ongoing issues within the model merging and propose seven research directions that are worthy of further exploration. We believe that model merging technology, as an efficient and modular model empowerment solution, will play a significant role in more practical scenarios in the future.
\bibliographystyle{ACM-Reference-Format}
\bibliography{main}

\end{document}